\documentclass[11pt]{article}

\usepackage[final]{acl}

\usepackage{times}
\usepackage{latexsym}

\usepackage[T1]{fontenc}

\usepackage[utf8]{inputenc}

\usepackage{microtype}

\usepackage{inconsolata}

\usepackage{float}
\usepackage{graphicx}
\usepackage{tabularx}

\usepackage{todonotes}
\usepackage{csquotes}
\usepackage{amsmath}
\usepackage{colortbl}
\usepackage{multirow}
\usepackage{booktabs}
\usepackage[shortlabels]{enumitem}
\usepackage{subcaption}
\usepackage[dvipsnames]{xcolor}
\usepackage{pgfplots}
\usetikzlibrary{math}
\usepgfplotslibrary{groupplots}
\usepackage{pgfplotstable}
\pgfplotsset{compat=1.18}
\usepackage{pgf-pie}
\usepackage[normalem]{ulem}

\definecolor{cA}{HTML}{4C72B0}   
\definecolor{cB}{HTML}{C44E52}   
\definecolor{cC}{HTML}{CCB742}   
\definecolor{cD}{HTML}{2D8E47}   
 
\newcommand{\boxwhisker}[8]{%
  \edef\xc{\the\numexpr#2\relax}
  \pgfmathsetmacro{\xpos}{#2+#8}%
  \draw[#1, thick] (axis cs:\xpos,#6) -- (axis cs:\xpos,#7);
  \pgfmathsetmacro{\capl}{\xpos-0.12}%
  \pgfmathsetmacro{\capr}{\xpos+0.12}%
  \draw[#1, thick] (axis cs:\capl,#6) -- (axis cs:\capr,#6);
  \draw[#1, thick] (axis cs:\capl,#7) -- (axis cs:\capr,#7);
  \pgfmathsetmacro{\boxl}{\xpos-0.18}%
  \pgfmathsetmacro{\boxr}{\xpos+0.18}%
  \fill[#1, opacity=0.25]
    (axis cs:\boxl,#4) rectangle (axis cs:\boxr,#5);
  \draw[#1, thick]
    (axis cs:\boxl,#4) rectangle (axis cs:\boxr,#5);
  \pgfmathsetmacro{\dashl}{\xpos-0.22}%
  \pgfmathsetmacro{\dashr}{\xpos+0.22}%
  \draw[#1, very thick] (axis cs:\dashl,#3) -- (axis cs:\dashr,#3);
}

\title{Research Design Tracking and Assessment for the Social Sciences}

\author{
  \textbf{Marco Rovera, Sergiu Burlacu, Dominique Cappelletti, Alessio Tomelleri,} \\
  \textbf{Sonia Marzadro, Martina Bazzoli, Annalisa Tassi, Jessica Gagete-Miranda} \\
  Fondazione Bruno Kessler \\
  Trento, Italy \\
  \texttt{m.rovera@fbk.eu} \\
  \texttt{\{sburlacu, dcappelletti, atomelleri, marzadro,} \\
  \texttt{bazzoli, atassi, jgagetemiranda\}@irvapp.it}
}

\begin{document}
\maketitle
\begin{abstract}

Reliable assessment of causal research designs in the social sciences is critical for evidence-based policy-making, yet has so far relied entirely on manual expert analysis. We introduce Automated Research Design Tracking and Assessment (ARDTrA), a task that involves detecting the research design used in a paper and assessing the quality of its application. We create an expert-annotated dataset of papers covering six families of \textit{counterfactual} research designs and evaluate the task using a multi-turn RAG-based conversational pipeline. Across four retrieval strategies, four LLMs and six embedding models, we find that passage length is the main driver of performance, explaining 52–66\% of the variance. A per-research-design analysis also shows that human and machine difficulty do not align: the designs that prove hardest for the system are not those on which expert annotators disagree most, pointing to two independent sources of task difficulty.
\end{abstract}

\section{Introduction}
\label{sec:intro}
Identifying and assessing the causal research designs employed in social science papers is a key yet underexplored NLP task. The demand for such automation is driven by a concrete use case: when policy advisors survey the scientific literature to inform decision-making on a specific topic, they need to quickly isolate studies whose methodology supports credible causal claims. In the social sciences, this translates into selecting studies that adopt a \textit{counterfactual} research design (RD), as for causal effect estimation, contributions employing such designs are generally regarded as providing more reliable evidence. Furthermore, even among papers that adopt a counterfactual RD, there are important differences in how the method is applied, which can affect the quality and robustness of the resulting evidence.
Due to the specialist knowledge required, regarding both RDs and their application caveats, this literature selection and assessment process has so far been carried out manually by social scientists. Despite its practical relevance, the task has received limited attention within the NLP community, likely due to the lack of both a formalized analytical framework and a public, expert-curated evaluation dataset.
In this paper we address these gaps by introducing Automated Research Design Tracking and Assessment (ARDTrA). The task involves two steps: (1) detecting which Research Design a paper employs, and (2) assessing the credibility, validity, and robustness of its application in the specific application context. Both steps require locating and interpreting fine-grained methodological details scattered throughout a paper.
Using an evaluation framework designed by experts in counterfactual policy evaluation, we first create and manually annotate an original corpus of 140 scientific papers drawn from a diverse range of social science fields, spanning applied economics, sociology, political science, and related disciplines.
The framework considers six families of RDs identified by the causal inference literature \citep{rubin1974estimating, Holland01121986} and subsequently consolidated by the so-called ``credibility revolution'' \citep{angrist2010credibility}, and develops an \textit{ad hoc} analytical scheme for each one. We then use the annotated dataset to evaluate different retrieval strategies within a multi-turn conversational RAG pipeline, across a set of LLMs and embedding models. From an NLP perspective, our primary goal is to evaluate the performance of various indexing and retrieval strategies on a challenging domain-specific document analysis task, where answers require synthesizing fine-grained methodological information from long scientific texts. From a Computational Social Science perspective, the study additionally demonstrates the feasibility of automating RD assessment and provides social scientists with a reliable estimate of current system performance.
The contribution of this paper is threefold: \textit{i)} we define the task of Research Design Tracking and Assessment, along with an expert-designed analytical schema; \textit{ii)} we create a dataset of 140 papers annotated by domain experts according to this schema; and \textit{iii)} we benchmark different retrieval strategies within a multi-turn RAG-based conversational agent, providing a systematic comparison across chunking methods, embedding models, and LLMs.

\section{Related Work}
\label{sec:related}
In the last few years, a growing body of literature has focused on automatic analysis of causal inference methods in the social sciences. \citet{currie2020technology} used basic text-mining techniques to track the incidence of keywords associated with causal inference methods in NBER\footnote{\url{https://www.nber.org/}} working papers and five leading economics journals. \citet{goldsmith2024tracking} extended this analysis to additional fields including finance and macroeconomics. More recently, \citet{garg2025causal} employed GPT-4o-mini to extract causal inference methods from NBER and CEPR\footnote{\url{https://www.cepr.org/}} working paper series, validating their approach against the annotated dataset of \citet{brodeur2024p}, which covers four RDs (DiD, RCT, RDD, IV) across 1,106 economics articles. While reporting accuracy scores between 70\% and 93\% depending on the RD, this study uses a single LLM without systematic evaluation of retrieval configurations, and focuses exclusively on RD \textit{detection} without addressing the quality of the method's application. Other related efforts include \citet{hooper2024semi}, who combine NLP with causal mapping for extracting causal evidence from policy literature, and \citet{imai2026causal}, who address causal inference methodology from a complementary perspective. OpenScholar \citep{asai2026synthesizing} used RAG on
45 million open-access papers, showing that 
retrieval-augmented systems substantially outperform parametric LLMs on scientific synthesis tasks. Our work differs in two ways: we introduce the \textit{assessment} dimension, i.e we evaluate not only which RD a paper uses but how \textit{credibly} it is applied, and we provide a systematic NLP evaluation across retrieval strategies, embedding models, and LLMs.

\section{Counterfactual Research Designs}
\label{sec:counterfactual_research_design}
In the social sciences, the credibility of a causal claim depends not only on the method employed but also on the specific empirical setting in which it is applied, as the \textit{causal identification} strategy plays a major role. For instance, in Card and Krueger's study of the employment effects of the minimum wage  \citep{card1993minimum}, the Difference-in-Differences design is credible precisely because it exploits a specific institutional contrast: New Jersey increased its minimum wage while neighboring Pennsylvania did not, providing a context-specific comparison group.
Causal claims are typically supported through research designs that attempt to construct or approximate such a comparison group, one that estimates the outcome that would have been observed in the absence of the intervention or treatment under study. Rather than relying exclusively on statistical associations, these approaches aim to exploit sources of variation that can plausibly isolate causal effects. 
Although the boundaries between causal and non-causal empirical strategies are not always sharp, the literature associated with the ``credibility revolution'' has identified a relatively consolidated set of design-based approaches that are widely regarded as suitable for causal inference in both experimental and observational settings \citep{imbens2024causal}. In this paper, we operationalize this literature by focusing on six broad families of research designs that cover the vast majority of causal designs encountered in applied social science research:

\begin{enumerate}
\item Experimental Designs (ED), e.g.\ field experiments, survey experiments and lab experiments \citep{fisher1966design, athey2017econometrics};
\item Threshold-based Designs (TbD), e.g.\ regression discontinuity, regression kink or bunching \citep{thistlethwaite1960regression, cattaneo2022regression};
\item Instrumental Variables (IV) \citep{angrist1996identification, mogstad2018identification};
\item Selection-on-Observables (SoO), e.g.\ regression adjustment, matching, weighting, doubly robust \citep{rubin1974estimating, heckman1979sample, stuart2010matching, imbens2015matching};
\item Difference-in-Differences (DiD) and related designs, e.g., event studies, changes-in-changes and triple differences \citep{Ashenfelter1985UsingTL, roth2023s};
\item Synthetic Control Methods (SCM) and related designs, e.g.\ augmented synthetic control, synthetic difference-in-differences \citep{abadie2003economic, abadie2010synthetic, abadie2021using}.
\end{enumerate}

It is worth noting that the analytical framework proposed in this work (see Section \ref{sec:dataset:guidelines} and Appendix \ref{sec:appendix_ardtra_fulltext}) is not specific to economics: many of its elements originate in, and are widely shared across, the empirical sciences as part of a common causal-inference toolkit \citep{imbens2015causal, hernan2020causal}. Experimental Designs, for instance, are foundational in psychology and medicine. Selection-on-Observables methods are central to epidemiology and public health, while other research designs, such as Difference-in-Differences, are increasingly adopted in related fields, like health policy \cite{feng2025difference}. Beyond academia, the same designs increasingly underpin large-scale experimentation and observational causal analysis in industry, particularly at large technology firms \cite{larsen2024statistical}. Collaborations between biostatisticians, econometricians and methodological experts in political science and sociology are common, and cross-referencing across these literatures is the norm in published work. While some differences exist, terminology is largely shared across disciplines, and our annotation scheme explicitly accounts for differences in use of certain terms (for example, \enquote{unconfoundedness}, \enquote{ignorability}, and \enquote{conditional independence assumption}, which refer to the same core assumption in Selection-on-Observables, are favored in different disciplines).

\section{Dataset}
\label{sec:dataset}
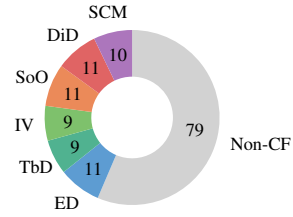
\begin{figure}[h]
\centering
\resizebox{0.5\columnwidth}{!}{%
\begin{tikzpicture}
\definecolor{cGrey}{RGB}{210,210,210}
\definecolor{c1}{RGB}{93,156,210}
\definecolor{c2}{RGB}{80,178,144}
\definecolor{c3}{RGB}{126,193,108}
\definecolor{c4}{RGB}{235,138,89}
\definecolor{c5}{RGB}{224,100,100}
\definecolor{c6}{RGB}{172,118,188}

\def\data{79/Non-CF/cGrey,
           11/ED/c1,
           9/TbD/c2,
           9/IV/c3,
           11/SoO/c4,
           11/DiD/c5,
           10/SCM/c6}
\def\total{140}
\def\innerR{0.7}
\def\outerR{1.5}
\pgfmathsetmacro{\currentangle}{90}
\foreach \val/\label/\col in \data {
    \pgfmathsetmacro{\sliceangle}{\val/\total*360}
    \pgfmathsetmacro{\endangle}{\currentangle - \sliceangle}
    \pgfmathsetmacro{\midangle}{(\currentangle + \endangle)/2}
    \pgfmathsetmacro{\midR}{(\innerR + \outerR)/2}
    \fill[\col] (\currentangle:\innerR) arc (\currentangle:\endangle:\innerR)
        -- (\endangle:\outerR) arc (\endangle:\currentangle:\outerR) -- cycle;
    \node[font=\small, text=black] at (\midangle:\midR) {\val};
    \pgfmathsetmacro{\labelR}{\outerR + (\val > 50 ? 0.75 : 0.35)}
    \node[font=\small] at (\midangle:\labelR) {\label};
    \xdef\currentangle{\endangle}
}
\end{tikzpicture}%
}
\caption{Distribution of RDs in the ARDTrA dataset.}
\label{fig:data_distribution}
\end{figure}

The dataset\footnote{Due to copyright restrictions, as most papers in the dataset are not open access, we release only the paper metadata (authors, title, journal) in the Supplementary Materials, along with the annotations and the analytical framework.} consists of 140 papers in English, sampled from an initial pool of 6,554 articles drawn from journals selected using the Combes-Linnemer classification \citep{combes2010inferring} and IDEAS/RePEc rankings, spanning multiple quality tiers and subfields, such as labour, health, education, environmental, urban and regional economics and was designed to maintain a reasonable balance between papers that use counterfactual and non-counterfactual RDs on the one hand, and between the various RDs on the other.
The distribution of RDs across the dataset is depicted in Figure \ref{fig:data_distribution}.

\subsection{Analytical Framework}
\label{sec:dataset:guidelines}
The analytical schema developed for ARDTrA is used for the manual annotation of papers in the dataset. The full text is provided in Appendix \ref{sec:appendix_ardtra_fulltext}. The schema consists of a set of methodological questions, each provided with instructions for answering (one or more valid answers) and with a finite set of possible answer options. The dataset was annotated by 8 domain experts with specialized knowledge of causal inference methodology and research design application. The task of each annotator is to read through the paper and answer the full set of questions. The schema is divided into two parts, which define the two subtasks: RD Identification and RD Assessment.

\paragraph{RD Identification} This subtask comprises 6 questions, arranged from general to specific, that progressively narrow down the methodological profile of a paper. The first three questions characterize the paper along broad dimensions: its type (e.g.\ empirical, theoretical, methodological, review), the nature of its data and research design (quantitative vs.\ qualitative), and the type of analysis it performs (causal, descriptive, or predictive/simulation). Question 4 then asks whether the paper employs any of the six counterfactual RDs introduced in Section \ref{sec:counterfactual_research_design}, as a binary yes/no decision. If the answer is negative, the paper is classified as non-counterfactual and the analysis terminates. If a counterfactual design is detected, Question 5 asks the annotator to identify which of the six families constitutes the paper's main RD. Finally, Question 6 captures whether the paper employs any additional counterfactual RDs beyond the primary one, allowing for multi-method designs. Papers identified as counterfactual at this stage proceed to the RD Assessment subtask.

\paragraph{RD Assessment} Once a counterfactual RD has been identified, the analysis proceeds to an assessment subtask tailored to the specific design identified in Question 5. Each of the six RD families has its own set of questions, ranging from 9 (Selection-on-Observables) to 16 (Instrumental Variables and Synthetic Control Methods), for a total of 76 questions across all designs. While the specific questions differ, reflecting the distinct assumptions and methodological concerns of each RD, the assessment schemes share a common structure organized around four broad dimensions: (a) the study context and specific features of the research design, (b) the features of the data available in the study, (c)   the estimation and inference procedures employed, and (d) falsification tests and and sensitivity analyses. For instance, the TbD schema asks about the features of the running variable, bandwidth selection and specific robustness checks, while the DiD schema focuses on the treatment assignment structure (blocked vs.\ staggered), parallel trends evidence, and the choice of estimator. All questions are closed-ended, with predefined answer options, and most allow multiple selections.

\subsection{Inter-Annotator Agreement}
\label{sec:dataset:iaa}

\begin{table}[t]
    \centering
    \small
    \begin{tabular}{l|c}
    \toprule
    & \textbf{Krippendorff's $\alpha$} \\
    \midrule
    Full Task & .814\\
    RD Identification & .926\\
    RD Assessment & .753\\
    \midrule
    Experimental Designs (ED) & .821 \\
    Synthetic Control Methods (SCM) & .833 \\
    Threshold-based Desings (TbD) & .784 \\
    Difference-in-Differences (DiD) & .688 \\
    Instrumental Variables (IV) & .681 \\
    Selection-on-Observables (SoO) & .649 \\
    \bottomrule
    \end{tabular}
    \caption{Krippendorff's $\alpha$ between 3 expert annotators, computed on the I-AA set (24 papers). Specific RD's scores have been computed on the 12 papers using a counterfactual RD (2 papers for RD).}
    \label{tab:iaa_scores}
\end{table}

\begin{table}[t]
\centering
\small
\begin{tabular}{@{}lccc@{}}
\toprule
\textbf{Annotator pair} & \textbf{Full Task} & \textbf{RD-I} & \textbf{RD-A} \\
\midrule
A1 - A2 & .820 & .922 & .767 \\
A1 - A3 & .771 & .925 & .686 \\
A2 - A3 & .851 & .935 & .805 \\
\bottomrule
\end{tabular}
\caption[Pairwise inter-annotator agreement]{Pairwise Krippendorff's $\alpha$ across the three tasks. A1, A2, A3 denote the three annotators.}
\label{tab:pairwise_iaa}
\end{table}

In order to estimate the complexity of the task, we asked three experts to independently annotate 24 scientific papers and computed inter-annotator agreement (I-AA) scores. The I-AA set consists of 12 papers using counterfactual methods (2 papers for each method) and 12 papers using non-counterfactual methods. Results are reported in Table \ref{tab:iaa_scores}).
The overall agreement on the Full Task (alpha = .81) is well above the .67 threshold commonly considered acceptable for reliable annotation \citep{krippendorff2004reliability}. Agreement is notably higher for RD Identification (.92) than for RD Assessment (.75), confirming that detecting a research design is substantially easier than evaluating how credibly it is applied, even for trained experts. The pairwise scores (Table \ref{tab:pairwise_iaa}) support this observation: RD Identification is uniformly high and stable across all annotator pairs (.92–.94), whereas RD Assessment is both lower and more variable, consistent with its more subjective nature. At the per-RD level, agreement varies considerably, ranging from ED and SCM at the top (.83 and .82) down to SoO (.64), IV (.68) and DiD (.69).

\section{Methodology}
\label{sec:experimental}
We frame the ARDTrA task as a RAG-based \citep{lewis2020retrieval, guu2020retrieval} multi-turn conversation with closed-ended questions. In this setup, the system processes one question at a time, in sequence: at each turn, the LLM receives (a) a question $q$, (b) its predefined answer options $a_1, a_2, ... a_n \in A$ and (c) a list of \textit{k} passages $p_1, p_2, ... p_n \in P$ dynamically retrieved from the paper under examination. The question \textit{q} and the answer options \textit{A} are concatenated and used jointly as the retrieval query, since the answer options contain method-specific terminology that improves passage selection. The conversation is maintained across turns within a fixed context window of 20K tokens, so that the model can leverage its previous answers when responding to subsequent, more specific assessment questions.
This design provides two key advantages. First, it makes the model's answers traceable: because retrieval selects a small set of passages for each question, every answer can be verified against specific spans of the source paper, a form of explainability that full-document prompting does not enable. Second, and more importantly, it closely replicates the human annotation process: the model answers the exact same closed-ended questions, with the same predefined answer options, in the same order as the human annotators, making human-machine comparison immediate. We compare four retrieval strategies that vary along the dimensions of chunking granularity (fixed vs.\ variable) and retrieval method (sparse vs.\ dense), using four LLMs of different sizes and six state-of-the-art embedding models. The four strategies are described below\footnote{All strategies have been implemented using Llamaindex. For Adaptive Auto-merging Retrieval and for Propositional Topic-aware we used the AutoMergingRetriever and the TopicNodeParser classes, respectively.}.

\paragraph{Best Match 25 (BM25)} BM25 \cite{robertson2009probabilistic} is included due to its low computational cost and its established effectiveness as a retrieval baseline. At indexing time, documents are chunked at fixed lengths and indexed in a sparse, keyword-based index. At retrieval time, the system uses the Okapi BM25 scoring function.

\paragraph{Dense Retrieval} At indexing time, documents are chunked at fixed lengths and each chunk is represented as a dense vector embedding. At retrieval time, passages are ranked by cosine similarity between the query embedding and the chunk embeddings.

\paragraph{Adaptive Auto-merging Retrieval} Adaptive retrieval is based on hierarchical indexing. At indexing time, the document is chunked at different lengths (we used 1024, 512, 256 and 128 tokens respectively). During parsing, chunks from each layer are structured into a tree, by keeping child/parent relations, as well as sibling relations. All nodes, at all levels, are then stored, but only leaf nodes are embedded, using dense embeddings. At retrieval time, similarity search is performed against the leaf-node index to retrieve an initial candidate set. A bottom-up merging pass then consolidates these candidates: for each parent node, the ratio of retrieved children to total children is computed, and when this ratio exceeds a threshold $t$ (we experimented with $t \in \{0.3, 0.5, 0.7\}$), the children are replaced by the parent node. This check recurses upward through the tree until no further merges are triggered. Lower $t$ values promote merging into fewer, larger chunks, while higher values keep the output closer to leaf retrieval, resulting in shorter and more specific chunks.

\paragraph{Propositional Topic-aware Retrieval} In this strategy, the document is first split into paragraphs. Each paragraph is then decomposed by an LLM (we used \texttt{Llama-3.1-8B-Instruct}) and \textit{rewritten} into a set of propositions, each representing an atomic statement contained in the paragraph \cite{chen-etal-2024-dense}. Propositions are the result of splitting compound sentences and resolving extra-propositional pronoun co-reference so that they are interpretable independently. Once paragraphs are converted into propositions, a parser iterates over propositions and builds \enquote{topic-aware} chunks by choosing whether to merge a new proposition into the current chunk or to create a new chunk. This decision is made by the parser based on a similarity threshold $st$, computed as the cosine similarity between the current chunk and the new proposition. Higher $st$ values result in smaller, topically more coherent chunks, whereas lower values produce larger, more inclusive and potentially more heterogeneous chunks. We experiment with $st \in \{0.3, 0.5, 0.8\}$. Also, a sliding window of size $ws$ is used by the parser to limit the context used in the chunk definition process. A small value of the $ws$ parameter will make the parser more sensitive to local topic switch, while a larger value results in more stable and inclusive topic chunks. We experiment with $ws \in \{2, 5\}$.
It is worth noting that, unlike all other strategies, in this case the chunks produced are not original text but consist of variable-length sequences of rewritten atomic propositions.

\paragraph{Long-Context} As a non-retrieval baseline, we additionally evaluate a Long-Context (LC) setting in which the full text of the paper, truncated to a 32K-token window for comparability across models, is provided directly in the model's context window without any retrieval. The model answers the same closed-ended questions, in the same multi-turn conversational format, but instead of receiving dynamically retrieved passages, it has access to the entire document. This baseline tests whether retrieval is beneficial at all, or whether a sufficiently large context window makes it unnecessary.

\subsection{Experimental Setup} The described methodology is instantiated on six embedding models of varying size and architecture: \texttt{mxbai-embed-large-v1} (335M), \texttt{bge-large-en-v1.5} (335M; \citealp{chen-etal-2024-m3}), \texttt{e5-large-v2} (335M; \citealp{wang2022text}), \texttt{SFR-Embedding-Mistral} (7B; \citealp{SFRAIResearch2024}), \texttt{Qwen3-Embedding-8B} (8B; \citealp{qwen3embedding}), and a proprietary model, \texttt{Cohere-embed-english-v3.0}.
In the main phase, all experiments are run using two LLMs, \texttt{Llama-3.1-8B-Instruct} and \texttt{Qwen2.5-32B-Instruct}. The top $k=5$ passages are retrieved at each turn. For each combination of strategy, hyperparameter values, embedding model and LLM, a complete run is conducted on the full dataset (140 documents). 
Considering the hyperparameter and embedding model combination, we run a total of 82 rounds for each LLM (4 for BM25, 24 for Dense, 18 for Adaptive and 36 for Propositional), summing up to 164 experimental rounds.
Based on the evidence collected in the main experimental phase, the best-performing embedding model is selected and a further set of experiments is then run with a single embedding model, using \texttt{Llama-3.3-70B-Instruct} and \texttt{gpt-5.1}, in order to benchmark larger LLMs while mitigating computational costs.

\section{Evaluation and Results}
\label{sec:eval}

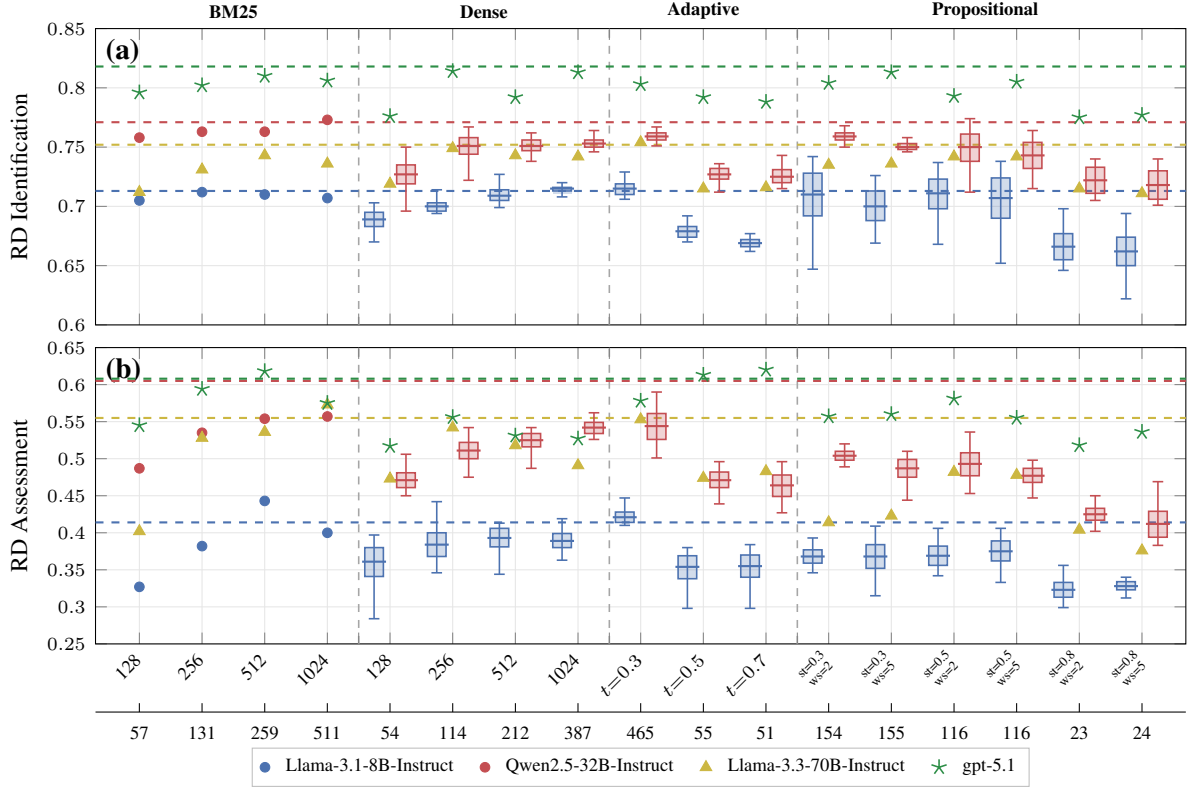
\begin{figure*}[t]
\centering
\begin{tikzpicture}
\begin{groupplot}[
  group style={
    group size=1 by 3,
    xlabels at=edge bottom,
    xticklabels at=edge bottom,
    vertical sep=0.3cm,
  },
  width=\textwidth,
  height=5.5cm,
  xmin=0.3, xmax=17.7,
  xtick={1,...,17},
  x tick label style={font=\scriptsize, rotate=45, anchor=east, yshift=-3pt},
  y tick label style={font=\scriptsize},
  ytick distance=0.05,
  grid=major,
  grid style={gray!20},
  extra x ticks={4.5, 8.5, 11.5},
  extra x tick labels={},
  extra x tick style={grid=major, grid style={black!35, dashed, semithick}},
  clip=false,
]

\nextgroupplot[
  ymin=0.6, ymax=0.85,
  ylabel={RD Identification},
  ylabel style={font=\small},
  xticklabels={},          
]
\node[font=\bfseries, anchor=north west]
  at (rel axis cs:0,1) {(a)};
\node[font=\scriptsize\bfseries, above] at (axis cs:2.5,0.85)  {BM25};
\node[font=\scriptsize\bfseries, above] at (axis cs:6.5,0.85)  {Dense};
\node[font=\scriptsize\bfseries, above] at (axis cs:10,0.85)   {Adaptive};
\node[font=\scriptsize\bfseries, above] at (axis cs:14.5,0.85) {Propositional};


\draw[cA, dashed, thick] (axis cs:0.3,0.713) -- (axis cs:17.7,0.713);
\draw[cB, dashed, thick] (axis cs:0.3,0.771) -- (axis cs:17.7,0.771);
\draw[cC, dashed, thick] (axis cs:0.3,0.752) -- (axis cs:17.7,0.752);
\draw[cD, dashed, thick] (axis cs:0.3,0.818) -- (axis cs:17.7,0.818);


\addplot[only marks, mark=*, mark size=1.8pt, color=cA]
  coordinates {(1,0.705)(2,0.712)(3,0.710)(4,0.707)};
\addplot[only marks, mark=*, mark size=1.8pt, color=cB]
  coordinates {(1,0.758)(2,0.763)(3,0.763)(4,0.773)};
\addplot[only marks, mark=triangle*, mark size=2.5pt, color=cC]
  coordinates {(1,0.712)(2,0.731)(3,0.743)(4,0.736)};
\addplot[only marks, mark=star, mark size=3pt, color=cD, semithick]
  coordinates {(1,0.796)(2,0.802)(3,0.810)(4,0.806)};
 
  \draw[cA, semithick] ([xshift=-6pt]axis cs:5,0.67) -- ([xshift=-6pt]axis cs:5,0.703);
  \draw[cA, semithick]
    ([xshift=-8pt]axis cs:5,0.67) -- ([xshift=-4pt]axis cs:5,0.67)
    ([xshift=-8pt]axis cs:5,0.703) -- ([xshift=-4pt]axis cs:5,0.703);
  \fill[cA, opacity=0.25]
    ([xshift=-9.5pt]axis cs:5,0.683) rectangle ([xshift=-2.5pt]axis cs:5,0.695);
  \draw[cA, semithick]
    ([xshift=-9.5pt]axis cs:5,0.683) rectangle ([xshift=-2.5pt]axis cs:5,0.695);
  \draw[cA, thick]
    ([xshift=-10.5pt]axis cs:5,0.689) -- ([xshift=-1.5pt]axis cs:5,0.689);
 
  \draw[cA, semithick] ([xshift=-6pt]axis cs:6,0.694) -- ([xshift=-6pt]axis cs:6,0.714);
  \draw[cA, semithick]
    ([xshift=-8pt]axis cs:6,0.694) -- ([xshift=-4pt]axis cs:6,0.694)
    ([xshift=-8pt]axis cs:6,0.714) -- ([xshift=-4pt]axis cs:6,0.714);
  \fill[cA, opacity=0.25]
    ([xshift=-9.5pt]axis cs:6,0.696) rectangle ([xshift=-2.5pt]axis cs:6,0.703);
  \draw[cA, semithick]
    ([xshift=-9.5pt]axis cs:6,0.696) rectangle ([xshift=-2.5pt]axis cs:6,0.703);
  \draw[cA, thick]
    ([xshift=-10.5pt]axis cs:6,0.700) -- ([xshift=-1.5pt]axis cs:6,0.700);
 
  \draw[cA, semithick] ([xshift=-6pt]axis cs:7,0.699) -- ([xshift=-6pt]axis cs:7,0.727);
  \draw[cA, semithick]
    ([xshift=-8pt]axis cs:7,0.699) -- ([xshift=-4pt]axis cs:7,0.699)
    ([xshift=-8pt]axis cs:7,0.727) -- ([xshift=-4pt]axis cs:7,0.727);
  \fill[cA, opacity=0.25]
    ([xshift=-9.5pt]axis cs:7,0.705) rectangle ([xshift=-2.5pt]axis cs:7,0.714);
  \draw[cA, semithick]
    ([xshift=-9.5pt]axis cs:7,0.705) rectangle ([xshift=-2.5pt]axis cs:7,0.714);
  \draw[cA, thick]
    ([xshift=-10.5pt]axis cs:7,0.709) -- ([xshift=-1.5pt]axis cs:7,0.709);
 
  \draw[cA, semithick] ([xshift=-6pt]axis cs:8,0.708) -- ([xshift=-6pt]axis cs:8,0.72);
  \draw[cA, semithick]
    ([xshift=-8pt]axis cs:8,0.708) -- ([xshift=-4pt]axis cs:8,0.708)
    ([xshift=-8pt]axis cs:8,0.72) -- ([xshift=-4pt]axis cs:8,0.72);
  \fill[cA, opacity=0.25]
    ([xshift=-9.5pt]axis cs:8,0.7103) rectangle ([xshift=-2.5pt]axis cs:8,0.716);
  \draw[cA, semithick]
    ([xshift=-9.5pt]axis cs:8,0.713) rectangle ([xshift=-2.5pt]axis cs:8,0.716);
  \draw[cA, thick]
    ([xshift=-10.5pt]axis cs:8,0.715) -- ([xshift=-1.5pt]axis cs:8,0.715);
 
  \draw[cA, semithick] ([xshift=-6pt]axis cs:9,0.706) -- ([xshift=-6pt]axis cs:9,0.729);
  \draw[cA, semithick]
    ([xshift=-8pt]axis cs:9,0.706) -- ([xshift=-4pt]axis cs:9,0.706)
    ([xshift=-8pt]axis cs:9,0.729) -- ([xshift=-4pt]axis cs:9,0.729);
  \fill[cA, opacity=0.25]
    ([xshift=-9.5pt]axis cs:9,0.710) rectangle ([xshift=-2.5pt]axis cs:9,0.719);
  \draw[cA, semithick]
    ([xshift=-9.5pt]axis cs:9,0.710) rectangle ([xshift=-2.5pt]axis cs:9,0.719);
  \draw[cA, thick]
    ([xshift=-10.5pt]axis cs:9,0.715) -- ([xshift=-1.5pt]axis cs:9,0.715);
 
  \draw[cA, semithick] ([xshift=-6pt]axis cs:10,0.67) -- ([xshift=-6pt]axis cs:10,0.692);
  \draw[cA, semithick]
    ([xshift=-8pt]axis cs:10,0.67) -- ([xshift=-4pt]axis cs:10,0.67)
    ([xshift=-8pt]axis cs:10,0.692) -- ([xshift=-4pt]axis cs:10,0.692);
  \fill[cA, opacity=0.25]
    ([xshift=-9.5pt]axis cs:10,0.674) rectangle ([xshift=-2.5pt]axis cs:10,0.683);
  \draw[cA, semithick]
    ([xshift=-9.5pt]axis cs:10,0.674) rectangle ([xshift=-2.5pt]axis cs:10,0.683);
  \draw[cA, thick]
    ([xshift=-10.5pt]axis cs:10,0.679) -- ([xshift=-1.5pt]axis cs:10,0.679);
 
  \draw[cA, semithick] ([xshift=-6pt]axis cs:11,0.662) -- ([xshift=-6pt]axis cs:11,0.677);
  \draw[cA, semithick]
    ([xshift=-8pt]axis cs:11,0.662) -- ([xshift=-4pt]axis cs:11,0.662)
    ([xshift=-8pt]axis cs:11,0.677) -- ([xshift=-4pt]axis cs:11,0.677);
  \fill[cA, opacity=0.25]
    ([xshift=-9.5pt]axis cs:11,0.666) rectangle ([xshift=-2.5pt]axis cs:11,0.672);
  \draw[cA, semithick]
    ([xshift=-9.5pt]axis cs:11,0.666) rectangle ([xshift=-2.5pt]axis cs:11,0.672);
  \draw[cA, thick]
    ([xshift=-10.5pt]axis cs:11,0.669) -- ([xshift=-1.5pt]axis cs:11,0.669);
 
  \draw[cA, semithick] ([xshift=-6pt]axis cs:12,0.647) -- ([xshift=-6pt]axis cs:12,0.742);
  \draw[cA, semithick]
    ([xshift=-8pt]axis cs:12,0.647) -- ([xshift=-4pt]axis cs:12,0.647)
    ([xshift=-8pt]axis cs:12,0.742) -- ([xshift=-4pt]axis cs:12,0.742);
  \fill[cA, opacity=0.25]
    ([xshift=-9.5pt]axis cs:12,0.692) rectangle ([xshift=-2.5pt]axis cs:12,0.728);
  \draw[cA, semithick]
    ([xshift=-9.5pt]axis cs:12,0.692) rectangle ([xshift=-2.5pt]axis cs:12,0.728);
  \draw[cA, thick]
    ([xshift=-10.5pt]axis cs:12,0.71) -- ([xshift=-1.5pt]axis cs:12,0.71);
 
  \draw[cA, semithick] ([xshift=-6pt]axis cs:13,0.669) -- ([xshift=-6pt]axis cs:13,0.726);
  \draw[cA, semithick]
    ([xshift=-8pt]axis cs:13,0.669) -- ([xshift=-4pt]axis cs:13,0.669)
    ([xshift=-8pt]axis cs:13,0.726) -- ([xshift=-4pt]axis cs:13,0.726);
  \fill[cA, opacity=0.25]
    ([xshift=-9.5pt]axis cs:13,0.688) rectangle ([xshift=-2.5pt]axis cs:13,0.713);
  \draw[cA, semithick]
    ([xshift=-9.5pt]axis cs:13,0.688) rectangle ([xshift=-2.5pt]axis cs:13,0.713);
  \draw[cA, thick]
    ([xshift=-10.5pt]axis cs:13,0.700) -- ([xshift=-1.5pt]axis cs:13,0.700);
 
  \draw[cA, semithick] ([xshift=-6pt]axis cs:14,0.668) -- ([xshift=-6pt]axis cs:14,0.737);
  \draw[cA, semithick]
    ([xshift=-8pt]axis cs:14,0.668) -- ([xshift=-4pt]axis cs:14,0.668)
    ([xshift=-8pt]axis cs:14,0.737) -- ([xshift=-4pt]axis cs:14,0.737);
  \fill[cA, opacity=0.25]
    ([xshift=-9.5pt]axis cs:14,0.698) rectangle ([xshift=-2.5pt]axis cs:14,0.723);
  \draw[cA, semithick]
    ([xshift=-9.5pt]axis cs:14,0.698) rectangle ([xshift=-2.5pt]axis cs:14,0.723);
  \draw[cA, thick]
    ([xshift=-10.5pt]axis cs:14,0.711) -- ([xshift=-1.5pt]axis cs:14,0.711);
 
  \draw[cA, semithick] ([xshift=-6pt]axis cs:15,0.652) -- ([xshift=-6pt]axis cs:15,0.738);
  \draw[cA, semithick]
    ([xshift=-8pt]axis cs:15,0.652) -- ([xshift=-4pt]axis cs:15,0.652)
    ([xshift=-8pt]axis cs:15,0.738) -- ([xshift=-4pt]axis cs:15,0.738);
  \fill[cA, opacity=0.25]
    ([xshift=-9.5pt]axis cs:15,0.690) rectangle ([xshift=-2.5pt]axis cs:15,0.724);
  \draw[cA, semithick]
    ([xshift=-9.5pt]axis cs:15,0.690) rectangle ([xshift=-2.5pt]axis cs:15,0.724);
  \draw[cA, thick]
    ([xshift=-10.5pt]axis cs:15,0.707) -- ([xshift=-1.5pt]axis cs:15,0.707);
 
  \draw[cA, semithick] ([xshift=-6pt]axis cs:16,0.646) -- ([xshift=-6pt]axis cs:16,0.698);
  \draw[cA, semithick]
    ([xshift=-8pt]axis cs:16,0.646) -- ([xshift=-4pt]axis cs:16,0.646)
    ([xshift=-8pt]axis cs:16,0.698) -- ([xshift=-4pt]axis cs:16,0.698);
  \fill[cA, opacity=0.25]
    ([xshift=-9.5pt]axis cs:16,0.655) rectangle ([xshift=-2.5pt]axis cs:16,0.677);
  \draw[cA, semithick]
    ([xshift=-9.5pt]axis cs:16,0.655) rectangle ([xshift=-2.5pt]axis cs:16,0.677);
  \draw[cA, thick]
    ([xshift=-10.5pt]axis cs:16,0.666) -- ([xshift=-1.5pt]axis cs:16,0.666);
 
  \draw[cA, semithick] ([xshift=-6pt]axis cs:17,0.622) -- ([xshift=-6pt]axis cs:17,0.694);
  \draw[cA, semithick]
    ([xshift=-8pt]axis cs:17,0.622) -- ([xshift=-4pt]axis cs:17,0.622)
    ([xshift=-8pt]axis cs:17,0.694) -- ([xshift=-4pt]axis cs:17,0.694);
  \fill[cA, opacity=0.25]
    ([xshift=-9.5pt]axis cs:17,0.650) rectangle ([xshift=-2.5pt]axis cs:17,0.674);
  \draw[cA, semithick]
    ([xshift=-9.5pt]axis cs:17,0.650) rectangle ([xshift=-2.5pt]axis cs:17,0.674);
  \draw[cA, thick]
    ([xshift=-10.5pt]axis cs:17,0.662) -- ([xshift=-1.5pt]axis cs:17,0.662);

  \draw[cB, semithick] ([xshift=6pt]axis cs:5,0.696) -- ([xshift=6pt]axis cs:5,0.75);
  \draw[cB, semithick]
    ([xshift=4pt]axis cs:5,0.696) -- ([xshift=8pt]axis cs:5,0.696)
    ([xshift=4pt]axis cs:5,0.75) -- ([xshift=8pt]axis cs:5,0.75);
  \fill[cB, opacity=0.25]
    ([xshift=2.5pt]axis cs:5,0.719) rectangle ([xshift=9.5pt]axis cs:5,0.735);
  \draw[cB, semithick]
    ([xshift=2.5pt]axis cs:5,0.719) rectangle ([xshift=9.5pt]axis cs:5,0.735);
  \draw[cB, thick]
    ([xshift=1.5pt]axis cs:5,0.727) -- ([xshift=10.5pt]axis cs:5,0.727);
 
  \draw[cB, semithick] ([xshift=6pt]axis cs:6,0.722) -- ([xshift=6pt]axis cs:6,0.767);
  \draw[cB, semithick]
    ([xshift=4pt]axis cs:6,0.722) -- ([xshift=8pt]axis cs:6,0.722)
    ([xshift=4pt]axis cs:6,0.767) -- ([xshift=8pt]axis cs:6,0.767);
  \fill[cB, opacity=0.25]
    ([xshift=2.5pt]axis cs:6,0.744) rectangle ([xshift=9.5pt]axis cs:6,0.758);
  \draw[cB, semithick]
    ([xshift=2.5pt]axis cs:6,0.744) rectangle ([xshift=9.5pt]axis cs:6,0.758);
  \draw[cB, thick]
    ([xshift=1.5pt]axis cs:6,0.751) -- ([xshift=10.5pt]axis cs:6,0.751);
 
  \draw[cB, semithick] ([xshift=6pt]axis cs:7,0.738) -- ([xshift=6pt]axis cs:7,0.762);
  \draw[cB, semithick]
    ([xshift=4pt]axis cs:7,0.738) -- ([xshift=8pt]axis cs:7,0.738)
    ([xshift=4pt]axis cs:7,0.762) -- ([xshift=8pt]axis cs:7,0.762);
  \fill[cB, opacity=0.25]
    ([xshift=2.5pt]axis cs:7,0.747) rectangle ([xshift=9.5pt]axis cs:7,0.756);
  \draw[cB, semithick]
    ([xshift=2.5pt]axis cs:7,0.747) rectangle ([xshift=9.5pt]axis cs:7,0.756);
  \draw[cB, thick]
    ([xshift=1.5pt]axis cs:7,0.751) -- ([xshift=10.5pt]axis cs:7,0.751);
 
  \draw[cB, semithick] ([xshift=6pt]axis cs:8,0.746) -- ([xshift=6pt]axis cs:8,0.764);
  \draw[cB, semithick]
    ([xshift=4pt]axis cs:8,0.746) -- ([xshift=8pt]axis cs:8,0.746)
    ([xshift=4pt]axis cs:8,0.764) -- ([xshift=8pt]axis cs:8,0.764);
  \fill[cB, opacity=0.25]
    ([xshift=2.5pt]axis cs:8,0.750) rectangle ([xshift=9.5pt]axis cs:8,0.756);
  \draw[cB, semithick]
    ([xshift=2.5pt]axis cs:8,0.750) rectangle ([xshift=9.5pt]axis cs:8,0.756);
  \draw[cB, thick]
    ([xshift=1.5pt]axis cs:8,0.753) -- ([xshift=10.5pt]axis cs:8,0.753);
 
  \draw[cB, semithick] ([xshift=6pt]axis cs:9,0.751) -- ([xshift=6pt]axis cs:9,0.767);
  \draw[cB, semithick]
    ([xshift=4pt]axis cs:9,0.751) -- ([xshift=8pt]axis cs:9,0.751)
    ([xshift=4pt]axis cs:9,0.767) -- ([xshift=8pt]axis cs:9,0.767);
  \fill[cB, opacity=0.25]
    ([xshift=2.5pt]axis cs:9,0.756) rectangle ([xshift=9.5pt]axis cs:9,0.762);
  \draw[cB, semithick]
    ([xshift=2.5pt]axis cs:9,0.756) rectangle ([xshift=9.5pt]axis cs:9,0.762);
  \draw[cB, thick]
    ([xshift=1.5pt]axis cs:9,0.759) -- ([xshift=10.5pt]axis cs:9,0.759);
 
  \draw[cB, semithick] ([xshift=6pt]axis cs:10,0.712) -- ([xshift=6pt]axis cs:10,0.736);
  \draw[cB, semithick]
    ([xshift=4pt]axis cs:10,0.712) -- ([xshift=8pt]axis cs:10,0.712)
    ([xshift=4pt]axis cs:10,0.736) -- ([xshift=8pt]axis cs:10,0.736);
  \fill[cB, opacity=0.25]
    ([xshift=2.5pt]axis cs:10,0.723) rectangle ([xshift=9.5pt]axis cs:10,0.732);
  \draw[cB, semithick]
    ([xshift=2.5pt]axis cs:10,0.723) rectangle ([xshift=9.5pt]axis cs:10,0.732);
  \draw[cB, thick]
    ([xshift=1.5pt]axis cs:10,0.727) -- ([xshift=10.5pt]axis cs:10,0.727);
 
  \draw[cB, semithick] ([xshift=6pt]axis cs:11,0.715) -- ([xshift=6pt]axis cs:11,0.743);
  \draw[cB, semithick]
    ([xshift=4pt]axis cs:11,0.715) -- ([xshift=8pt]axis cs:11,0.715)
    ([xshift=4pt]axis cs:11,0.743) -- ([xshift=8pt]axis cs:11,0.743);
  \fill[cB, opacity=0.25]
    ([xshift=2.5pt]axis cs:11,0.720) rectangle ([xshift=9.5pt]axis cs:11,0.731);
  \draw[cB, semithick]
    ([xshift=2.5pt]axis cs:11,0.720) rectangle ([xshift=9.5pt]axis cs:11,0.731);
  \draw[cB, thick]
    ([xshift=1.5pt]axis cs:11,0.725) -- ([xshift=10.5pt]axis cs:11,0.725);
 
  \draw[cB, semithick] ([xshift=6pt]axis cs:12,0.75) -- ([xshift=6pt]axis cs:12,0.768);
  \draw[cB, semithick]
    ([xshift=4pt]axis cs:12,0.75) -- ([xshift=8pt]axis cs:12,0.75)
    ([xshift=4pt]axis cs:12,0.768) -- ([xshift=8pt]axis cs:12,0.768);
  \fill[cB, opacity=0.25]
    ([xshift=2.5pt]axis cs:12,0.756) rectangle ([xshift=9.5pt]axis cs:12,0.762);
  \draw[cB, semithick]
    ([xshift=2.5pt]axis cs:12,0.756) rectangle ([xshift=9.5pt]axis cs:12,0.762);
  \draw[cB, thick]
    ([xshift=1.5pt]axis cs:12,0.759) -- ([xshift=10.5pt]axis cs:12,0.759);
 
  \draw[cB, semithick] ([xshift=6pt]axis cs:13,0.746) -- ([xshift=6pt]axis cs:13,0.758);
  \draw[cB, semithick]
    ([xshift=4pt]axis cs:13,0.746) -- ([xshift=8pt]axis cs:13,0.746)
    ([xshift=4pt]axis cs:13,0.758) -- ([xshift=8pt]axis cs:13,0.758);
  \fill[cB, opacity=0.25]
    ([xshift=2.5pt]axis cs:13,0.748) rectangle ([xshift=9.5pt]axis cs:13,0.753);
  \draw[cB, semithick]
    ([xshift=2.5pt]axis cs:13,0.748) rectangle ([xshift=9.5pt]axis cs:13,0.753);
  \draw[cB, thick]
    ([xshift=1.5pt]axis cs:13,0.750) -- ([xshift=10.5pt]axis cs:13,0.750);
 
  \draw[cB, semithick] ([xshift=6pt]axis cs:14,0.712) -- ([xshift=6pt]axis cs:14,0.774);
  \draw[cB, semithick]
    ([xshift=4pt]axis cs:14,0.712) -- ([xshift=8pt]axis cs:14,0.712)
    ([xshift=4pt]axis cs:14,0.774) -- ([xshift=8pt]axis cs:14,0.774);
  \fill[cB, opacity=0.25]
    ([xshift=2.5pt]axis cs:14,0.738) rectangle ([xshift=9.5pt]axis cs:14,0.761);
  \draw[cB, semithick]
    ([xshift=2.5pt]axis cs:14,0.738) rectangle ([xshift=9.5pt]axis cs:14,0.761);
  \draw[cB, thick]
    ([xshift=1.5pt]axis cs:14,0.750) -- ([xshift=10.5pt]axis cs:14,0.750);
 
  \draw[cB, semithick] ([xshift=6pt]axis cs:15,0.715) -- ([xshift=6pt]axis cs:15,0.764);
  \draw[cB, semithick]
    ([xshift=4pt]axis cs:15,0.715) -- ([xshift=8pt]axis cs:15,0.715)
    ([xshift=4pt]axis cs:15,0.764) -- ([xshift=8pt]axis cs:15,0.764);
  \fill[cB, opacity=0.25]
    ([xshift=2.5pt]axis cs:15,0.732) rectangle ([xshift=9.5pt]axis cs:15,0.754);
  \draw[cB, semithick]
    ([xshift=2.5pt]axis cs:15,0.732) rectangle ([xshift=9.5pt]axis cs:15,0.754);
  \draw[cB, thick]
    ([xshift=1.5pt]axis cs:15,0.743) -- ([xshift=10.5pt]axis cs:15,0.743);
 
  \draw[cB, semithick] ([xshift=6pt]axis cs:16,0.705) -- ([xshift=6pt]axis cs:16,0.74);
  \draw[cB, semithick]
    ([xshift=4pt]axis cs:16,0.705) -- ([xshift=8pt]axis cs:16,0.705)
    ([xshift=4pt]axis cs:16,0.74) -- ([xshift=8pt]axis cs:16,0.74);
  \fill[cB, opacity=0.25]
    ([xshift=2.5pt]axis cs:16,0.711) rectangle ([xshift=9.5pt]axis cs:16,0.733);
  \draw[cB, semithick]
    ([xshift=2.5pt]axis cs:16,0.711) rectangle ([xshift=9.5pt]axis cs:16,0.733);
  \draw[cB, thick]
    ([xshift=1.5pt]axis cs:16,0.722) -- ([xshift=10.5pt]axis cs:16,0.722);
 
  \draw[cB, semithick] ([xshift=6pt]axis cs:17,0.701) -- ([xshift=6pt]axis cs:17,0.74);
  \draw[cB, semithick]
    ([xshift=4pt]axis cs:17,0.701) -- ([xshift=8pt]axis cs:17,0.701)
    ([xshift=4pt]axis cs:17,0.74) -- ([xshift=8pt]axis cs:17,0.74);
  \fill[cB, opacity=0.25]
    ([xshift=2.5pt]axis cs:17,0.706) rectangle ([xshift=9.5pt]axis cs:17,0.730);
  \draw[cB, semithick]
    ([xshift=2.5pt]axis cs:17,0.706) rectangle ([xshift=9.5pt]axis cs:17,0.730);
  \draw[cB, thick]
    ([xshift=1.5pt]axis cs:17,0.718) -- ([xshift=10.5pt]axis cs:17,0.718);
 
\addplot[only marks, mark=triangle*, mark size=2.5pt, color=cC, forget plot]
  coordinates {
    (5,0.719)(6,0.749)(7,0.743)(8,0.742)
    (9,0.754)(10,0.715)(11,0.716)
    (12,0.735)(13,0.736)(14,0.742)(15,0.742)(16,0.715)(17,0.711)};
 
\addplot[only marks, mark=star, mark size=3pt, color=cD, semithick, forget plot]
  coordinates {
    (5,0.776)(6,0.814)(7,0.792)(8,0.813)
    (9,0.803)(10,0.792)(11,0.788)
    (12,0.804)(13,0.813)(14,0.793)(15,0.805)(16,0.775)(17,0.777)};

\nextgroupplot[
  ymin=0.25, ymax=0.65,
  ylabel={RD Assessment},
  ylabel style={font=\small},
  xticklabels={                           
    128, 256, 512, 1024,%
    128, 256, 512, 1024,%
    $t{=}0.3$, $t{=}0.5$, $t{=}0.7$,%
    {$\substack{\text{st=0.3}\\\text{ws=2}}$},
    {$\substack{\text{st=0.3}\\\text{ws=5}}$},
    {$\substack{\text{st=0.5}\\\text{ws=2}}$},
    {$\substack{\text{st=0.5}\\\text{ws=5}}$},
    {$\substack{\text{st=0.8}\\\text{ws=2}}$},
    {$\substack{\text{st=0.8}\\\text{ws=5}}$}},
  legend columns=4,
  legend style={
    at={(0.5,-0.35)}, anchor=north,
    font=\scriptsize, draw=gray!50,
    cells={anchor=west},
    column sep=4pt,
    /tikz/every even column/.append style={column sep=8pt},
  },
]
\node[font=\bfseries, anchor=north west]
  at (rel axis cs:0,1) {(b)};


\draw[cA, dashed, thick] (axis cs:0.3,0.414) -- (axis cs:17.7,0.414);
\draw[cB, dashed, thick] (axis cs:0.3,0.605) -- (axis cs:17.7,0.605);
\draw[cC, dashed, thick] (axis cs:0.3,0.555) -- (axis cs:17.7,0.555);
\draw[cD, dashed, thick] (axis cs:0.3,0.608) -- (axis cs:17.7,0.608);

\addplot[only marks, mark=*, mark size=1.8pt, color=cA]
  coordinates {(1,0.327)(2,0.382)(3,0.443)(4,0.4)};
\addplot[only marks, mark=*, mark size=1.8pt, color=cB]
  coordinates {(1,0.487)(2,0.535)(3,0.554)(4,0.557)};
\addplot[only marks, mark=triangle*, mark size=2.5pt, color=cC]
  coordinates {(1,0.402)(2,0.528)(3,0.536)(4,0.573)};
\addplot[only marks, mark=star, mark size=3pt, color=cD, semithick]
  coordinates {(1,0.545)(2,0.594)(3,0.618)(4,0.575)};
 
  \draw[cA, semithick] ([xshift=-6pt]axis cs:5,0.284) -- ([xshift=-6pt]axis cs:5,0.397);
  \draw[cA, semithick]
    ([xshift=-8pt]axis cs:5,0.284) -- ([xshift=-4pt]axis cs:5,0.284)
    ([xshift=-8pt]axis cs:5,0.397) -- ([xshift=-4pt]axis cs:5,0.397);
  \fill[cA, opacity=0.25]
    ([xshift=-9.5pt]axis cs:5,0.341) rectangle ([xshift=-2.5pt]axis cs:5,0.38);
  \draw[cA, semithick]
    ([xshift=-9.5pt]axis cs:5,0.341) rectangle ([xshift=-2.5pt]axis cs:5,0.380);
  \draw[cA, thick]
    ([xshift=-10.5pt]axis cs:5,0.361) -- ([xshift=-1.5pt]axis cs:5,0.361);
 
  \draw[cA, semithick] ([xshift=-6pt]axis cs:6,0.346) -- ([xshift=-6pt]axis cs:6,0.442);
  \draw[cA, semithick]
    ([xshift=-8pt]axis cs:6,0.346) -- ([xshift=-4pt]axis cs:6,0.346)
    ([xshift=-8pt]axis cs:6,0.442) -- ([xshift=-4pt]axis cs:6,0.442);
  \fill[cA, opacity=0.25]
    ([xshift=-9.5pt]axis cs:6,0.368) rectangle ([xshift=-2.5pt]axis cs:6,0.400);
  \draw[cA, semithick]
    ([xshift=-9.5pt]axis cs:6,0.368) rectangle ([xshift=-2.5pt]axis cs:6,0.400);
  \draw[cA, thick]
    ([xshift=-10.5pt]axis cs:6,0.384) -- ([xshift=-1.5pt]axis cs:6,0.384);
 
  \draw[cA, semithick] ([xshift=-6pt]axis cs:7,0.344) -- ([xshift=-6pt]axis cs:7,0.413);
  \draw[cA, semithick]
    ([xshift=-8pt]axis cs:7,0.344) -- ([xshift=-4pt]axis cs:7,0.344)
    ([xshift=-8pt]axis cs:7,0.413) -- ([xshift=-4pt]axis cs:7,0.413);
  \fill[cA, opacity=0.25]
    ([xshift=-9.5pt]axis cs:7,0.381) rectangle ([xshift=-2.5pt]axis cs:7,0.406);
  \draw[cA, semithick]
    ([xshift=-9.5pt]axis cs:7,0.381) rectangle ([xshift=-2.5pt]axis cs:7,0.406);
  \draw[cA, thick]
    ([xshift=-10.5pt]axis cs:7,0.393) -- ([xshift=-1.5pt]axis cs:7,0.393);
 
  \draw[cA, semithick] ([xshift=-6pt]axis cs:8,0.363) -- ([xshift=-6pt]axis cs:8,0.419);
  \draw[cA, semithick]
    ([xshift=-8pt]axis cs:8,0.363) -- ([xshift=-4pt]axis cs:8,0.363)
    ([xshift=-8pt]axis cs:8,0.419) -- ([xshift=-4pt]axis cs:8,0.419);
  \fill[cA, opacity=0.25]
    ([xshift=-9.5pt]axis cs:8,0.380) rectangle ([xshift=-2.5pt]axis cs:8,0.399);
  \draw[cA, semithick]
    ([xshift=-9.5pt]axis cs:8,0.380) rectangle ([xshift=-2.5pt]axis cs:8,0.399);
  \draw[cA, thick]
    ([xshift=-10.5pt]axis cs:8,0.389) -- ([xshift=-1.5pt]axis cs:8,0.389);
 
  \draw[cA, semithick] ([xshift=-6pt]axis cs:9,0.410) -- ([xshift=-6pt]axis cs:9,0.447);
  \draw[cA, semithick]
    ([xshift=-8pt]axis cs:9,0.410) -- ([xshift=-4pt]axis cs:9,0.410)
    ([xshift=-8pt]axis cs:9,0.447) -- ([xshift=-4pt]axis cs:9,0.447);
  \fill[cA, opacity=0.25]
    ([xshift=-9.5pt]axis cs:9,0.414) rectangle ([xshift=-2.5pt]axis cs:9,0.428);
  \draw[cA, semithick]
    ([xshift=-9.5pt]axis cs:9,0.414) rectangle ([xshift=-2.5pt]axis cs:9,0.428);
  \draw[cA, thick]
    ([xshift=-10.5pt]axis cs:9,0.421) -- ([xshift=-1.5pt]axis cs:9,0.421);
 
  \draw[cA, semithick] ([xshift=-6pt]axis cs:10,0.298) -- ([xshift=-6pt]axis cs:10,0.380);
  \draw[cA, semithick]
    ([xshift=-8pt]axis cs:10,0.298) -- ([xshift=-4pt]axis cs:10,0.298)
    ([xshift=-8pt]axis cs:10,0.380) -- ([xshift=-4pt]axis cs:10,0.380);
  \fill[cA, opacity=0.25]
    ([xshift=-9.5pt]axis cs:10,0.338) rectangle ([xshift=-2.5pt]axis cs:10,0.369);
  \draw[cA, semithick]
    ([xshift=-9.5pt]axis cs:10,0.338) rectangle ([xshift=-2.5pt]axis cs:10,0.369);
  \draw[cA, thick]
    ([xshift=-10.5pt]axis cs:10,0.354) -- ([xshift=-1.5pt]axis cs:10,0.354);
 
  \draw[cA, semithick] ([xshift=-6pt]axis cs:11,0.298) -- ([xshift=-6pt]axis cs:11,0.384);
  \draw[cA, semithick]
    ([xshift=-8pt]axis cs:11,0.298) -- ([xshift=-4pt]axis cs:11,0.298)
    ([xshift=-8pt]axis cs:11,0.384) -- ([xshift=-4pt]axis cs:11,0.384);
  \fill[cA, opacity=0.25]
    ([xshift=-9.5pt]axis cs:11,0.340) rectangle ([xshift=-2.5pt]axis cs:11,0.370);
  \draw[cA, semithick]
    ([xshift=-9.5pt]axis cs:11,0.340) rectangle ([xshift=-2.5pt]axis cs:11,0.370);
  \draw[cA, thick]
    ([xshift=-10.5pt]axis cs:11,0.355) -- ([xshift=-1.5pt]axis cs:11,0.355);
 
  \draw[cA, semithick] ([xshift=-6pt]axis cs:12,0.346) -- ([xshift=-6pt]axis cs:12,0.393);
  \draw[cA, semithick]
    ([xshift=-8pt]axis cs:12,0.346) -- ([xshift=-4pt]axis cs:12,0.346)
    ([xshift=-8pt]axis cs:12,0.393) -- ([xshift=-4pt]axis cs:12,0.393);
  \fill[cA, opacity=0.25]
    ([xshift=-9.5pt]axis cs:12,0.359) rectangle ([xshift=-2.5pt]axis cs:12,0.377);
  \draw[cA, semithick]
    ([xshift=-9.5pt]axis cs:12,0.359) rectangle ([xshift=-2.5pt]axis cs:12,0.377);
  \draw[cA, thick]
    ([xshift=-10.5pt]axis cs:12,0.368) -- ([xshift=-1.5pt]axis cs:12,0.368);
 
  \draw[cA, semithick] ([xshift=-6pt]axis cs:13,0.315) -- ([xshift=-6pt]axis cs:13,0.409);
  \draw[cA, semithick]
    ([xshift=-8pt]axis cs:13,0.315) -- ([xshift=-4pt]axis cs:13,0.315)
    ([xshift=-8pt]axis cs:13,0.409) -- ([xshift=-4pt]axis cs:13,0.409);
  \fill[cA, opacity=0.25]
    ([xshift=-9.5pt]axis cs:13,0.352) rectangle ([xshift=-2.5pt]axis cs:13,0.384);
  \draw[cA, semithick]
    ([xshift=-9.5pt]axis cs:13,0.352) rectangle ([xshift=-2.5pt]axis cs:13,0.384);
  \draw[cA, thick]
    ([xshift=-10.5pt]axis cs:13,0.368) -- ([xshift=-1.5pt]axis cs:13,0.368);
 
  \draw[cA, semithick] ([xshift=-6pt]axis cs:14,0.342) -- ([xshift=-6pt]axis cs:14,0.406);
  \draw[cA, semithick]
    ([xshift=-8pt]axis cs:14,0.342) -- ([xshift=-4pt]axis cs:14,0.342)
    ([xshift=-8pt]axis cs:14,0.406) -- ([xshift=-4pt]axis cs:14,0.406);
  \fill[cA, opacity=0.25]
    ([xshift=-9.5pt]axis cs:14,0.356) rectangle ([xshift=-2.5pt]axis cs:14,0.382);
  \draw[cA, semithick]
    ([xshift=-9.5pt]axis cs:14,0.356) rectangle ([xshift=-2.5pt]axis cs:14,0.382);
  \draw[cA, thick]
    ([xshift=-10.5pt]axis cs:14,0.369) -- ([xshift=-1.5pt]axis cs:14,0.369);
 
  \draw[cA, semithick] ([xshift=-6pt]axis cs:15,0.333) -- ([xshift=-6pt]axis cs:15,0.406);
  \draw[cA, semithick]
    ([xshift=-8pt]axis cs:15,0.333) -- ([xshift=-4pt]axis cs:15,0.333)
    ([xshift=-8pt]axis cs:15,0.406) -- ([xshift=-4pt]axis cs:15,0.406);
  \fill[cA, opacity=0.25]
    ([xshift=-9.5pt]axis cs:15,0.362) rectangle ([xshift=-2.5pt]axis cs:15,0.389);
  \draw[cA, semithick]
    ([xshift=-9.5pt]axis cs:15,0.362) rectangle ([xshift=-2.5pt]axis cs:15,0.389);
  \draw[cA, thick]
    ([xshift=-10.5pt]axis cs:15,0.375) -- ([xshift=-1.5pt]axis cs:15,0.375);
 
  \draw[cA, semithick] ([xshift=-6pt]axis cs:16,0.299) -- ([xshift=-6pt]axis cs:16,0.356);
  \draw[cA, semithick]
    ([xshift=-8pt]axis cs:16,0.299) -- ([xshift=-4pt]axis cs:16,0.299)
    ([xshift=-8pt]axis cs:16,0.356) -- ([xshift=-4pt]axis cs:16,0.356);
  \fill[cA, opacity=0.25]
    ([xshift=-9.5pt]axis cs:16,0.313) rectangle ([xshift=-2.5pt]axis cs:16,0.333);
  \draw[cA, semithick]
    ([xshift=-9.5pt]axis cs:16,0.313) rectangle ([xshift=-2.5pt]axis cs:16,0.333);
  \draw[cA, thick]
    ([xshift=-10.5pt]axis cs:16,0.323) -- ([xshift=-1.5pt]axis cs:16,0.323);
 
  \draw[cA, semithick] ([xshift=-6pt]axis cs:17,0.312) -- ([xshift=-6pt]axis cs:17,0.340);
  \draw[cA, semithick]
    ([xshift=-8pt]axis cs:17,0.312) -- ([xshift=-4pt]axis cs:17,0.312)
    ([xshift=-8pt]axis cs:17,0.340) -- ([xshift=-4pt]axis cs:17,0.340);
  \fill[cA, opacity=0.25]
    ([xshift=-9.5pt]axis cs:17,0.323) rectangle ([xshift=-2.5pt]axis cs:17,0.334);
  \draw[cA, semithick]
    ([xshift=-9.5pt]axis cs:17,0.323) rectangle ([xshift=-2.5pt]axis cs:17,0.334);
  \draw[cA, thick]
    ([xshift=-10.5pt]axis cs:17,0.328) -- ([xshift=-1.5pt]axis cs:17,0.328);

  \draw[cB, semithick] ([xshift=6pt]axis cs:5,0.450) -- ([xshift=6pt]axis cs:5,0.506);
  \draw[cB, semithick]
    ([xshift=4pt]axis cs:5,0.450) -- ([xshift=8pt]axis cs:5,0.450)
    ([xshift=4pt]axis cs:5,0.506) -- ([xshift=8pt]axis cs:5,0.506);
  \fill[cB, opacity=0.25]
    ([xshift=2.5pt]axis cs:5,0.461) rectangle ([xshift=9.5pt]axis cs:5,0.481);
  \draw[cB, semithick]
    ([xshift=2.5pt]axis cs:5,0.461) rectangle ([xshift=9.5pt]axis cs:5,0.481);
  \draw[cB, thick]
    ([xshift=1.5pt]axis cs:5,0.471) -- ([xshift=10.5pt]axis cs:5,0.471);
 
  \draw[cB, semithick] ([xshift=6pt]axis cs:6,0.475) -- ([xshift=6pt]axis cs:6,0.542);
  \draw[cB, semithick]
    ([xshift=4pt]axis cs:6,0.475) -- ([xshift=8pt]axis cs:6,0.475)
    ([xshift=4pt]axis cs:6,0.542) -- ([xshift=8pt]axis cs:6,0.542);
  \fill[cB, opacity=0.25]
    ([xshift=2.5pt]axis cs:6,0.500) rectangle ([xshift=9.5pt]axis cs:6,0.522);
  \draw[cB, semithick]
    ([xshift=2.5pt]axis cs:6,0.500) rectangle ([xshift=9.5pt]axis cs:6,0.522);
  \draw[cB, thick]
    ([xshift=1.5pt]axis cs:6,0.511) -- ([xshift=10.5pt]axis cs:6,0.511);
 
  \draw[cB, semithick] ([xshift=6pt]axis cs:7,0.487) -- ([xshift=6pt]axis cs:7,0.542);
  \draw[cB, semithick]
    ([xshift=4pt]axis cs:7,0.487) -- ([xshift=8pt]axis cs:7,0.487)
    ([xshift=4pt]axis cs:7,0.542) -- ([xshift=8pt]axis cs:7,0.542);
  \fill[cB, opacity=0.25]
    ([xshift=2.5pt]axis cs:7,0.516) rectangle ([xshift=9.5pt]axis cs:7,0.534);
  \draw[cB, semithick]
    ([xshift=2.5pt]axis cs:7,0.516) rectangle ([xshift=9.5pt]axis cs:7,0.534);
  \draw[cB, thick]
    ([xshift=1.5pt]axis cs:7,0.525) -- ([xshift=10.5pt]axis cs:7,0.525);
 
  \draw[cB, semithick] ([xshift=6pt]axis cs:8,0.526) -- ([xshift=6pt]axis cs:8,0.562);
  \draw[cB, semithick]
    ([xshift=4pt]axis cs:8,0.526) -- ([xshift=8pt]axis cs:8,0.526)
    ([xshift=4pt]axis cs:8,0.562) -- ([xshift=8pt]axis cs:8,0.562);
  \fill[cB, opacity=0.25]
    ([xshift=2.5pt]axis cs:8,0.534) rectangle ([xshift=9.5pt]axis cs:8,0.549);
  \draw[cB, semithick]
    ([xshift=2.5pt]axis cs:8,0.534) rectangle ([xshift=9.5pt]axis cs:8,0.549);
  \draw[cB, thick]
    ([xshift=1.5pt]axis cs:8,0.542) -- ([xshift=10.5pt]axis cs:8,0.542);
 
  \draw[cB, semithick] ([xshift=6pt]axis cs:9,0.501) -- ([xshift=6pt]axis cs:9,0.590);
  \draw[cB, semithick]
    ([xshift=4pt]axis cs:9,0.501) -- ([xshift=8pt]axis cs:9,0.501)
    ([xshift=4pt]axis cs:9,0.590) -- ([xshift=8pt]axis cs:9,0.590);
  \fill[cB, opacity=0.25]
    ([xshift=2.5pt]axis cs:9,0.526) rectangle ([xshift=9.5pt]axis cs:9,0.561);
  \draw[cB, semithick]
    ([xshift=2.5pt]axis cs:9,0.526) rectangle ([xshift=9.5pt]axis cs:9,0.561);
  \draw[cB, thick]
    ([xshift=1.5pt]axis cs:9,0.544) -- ([xshift=10.5pt]axis cs:9,0.544);
 
  \draw[cB, semithick] ([xshift=6pt]axis cs:10,0.439) -- ([xshift=6pt]axis cs:10,0.496);
  \draw[cB, semithick]
    ([xshift=4pt]axis cs:10,0.439) -- ([xshift=8pt]axis cs:10,0.439)
    ([xshift=4pt]axis cs:10,0.496) -- ([xshift=8pt]axis cs:10,0.496);
  \fill[cB, opacity=0.25]
    ([xshift=2.5pt]axis cs:10,0.461) rectangle ([xshift=9.5pt]axis cs:10,0.482);
  \draw[cB, semithick]
    ([xshift=2.5pt]axis cs:10,0.461) rectangle ([xshift=9.5pt]axis cs:10,0.482);
  \draw[cB, thick]
    ([xshift=1.5pt]axis cs:10,0.471) -- ([xshift=10.5pt]axis cs:10,0.471);
 
  \draw[cB, semithick] ([xshift=6pt]axis cs:11,0.427) -- ([xshift=6pt]axis cs:11,0.496);
  \draw[cB, semithick]
    ([xshift=4pt]axis cs:11,0.427) -- ([xshift=8pt]axis cs:11,0.427)
    ([xshift=4pt]axis cs:11,0.496) -- ([xshift=8pt]axis cs:11,0.496);
  \fill[cB, opacity=0.25]
    ([xshift=2.5pt]axis cs:11,0.449) rectangle ([xshift=9.5pt]axis cs:11,0.478);
  \draw[cB, semithick]
    ([xshift=2.5pt]axis cs:11,0.449) rectangle ([xshift=9.5pt]axis cs:11,0.478);
  \draw[cB, thick]
    ([xshift=1.5pt]axis cs:11,0.464) -- ([xshift=10.5pt]axis cs:11,0.464);
 
  \draw[cB, semithick] ([xshift=6pt]axis cs:12,0.489) -- ([xshift=6pt]axis cs:12,0.52);
  \draw[cB, semithick]
    ([xshift=4pt]axis cs:12,0.489) -- ([xshift=8pt]axis cs:12,0.489)
    ([xshift=4pt]axis cs:12,0.520) -- ([xshift=8pt]axis cs:12,0.520);
  \fill[cB, opacity=0.25]
    ([xshift=2.5pt]axis cs:12,0.498) rectangle ([xshift=9.5pt]axis cs:12,0.510);
  \draw[cB, semithick]
    ([xshift=2.5pt]axis cs:12,0.498) rectangle ([xshift=9.5pt]axis cs:12,0.510);
  \draw[cB, thick]
    ([xshift=1.5pt]axis cs:12,0.504) -- ([xshift=10.5pt]axis cs:12,0.504);
 
  \draw[cB, semithick] ([xshift=6pt]axis cs:13,0.444) -- ([xshift=6pt]axis cs:13,0.510);
  \draw[cB, semithick]
    ([xshift=4pt]axis cs:13,0.444) -- ([xshift=8pt]axis cs:13,0.444)
    ([xshift=4pt]axis cs:13,0.510) -- ([xshift=8pt]axis cs:13,0.510);
  \fill[cB, opacity=0.25]
    ([xshift=2.5pt]axis cs:13,0.475) rectangle ([xshift=9.5pt]axis cs:13,0.499);
  \draw[cB, semithick]
    ([xshift=2.5pt]axis cs:13,0.475) rectangle ([xshift=9.5pt]axis cs:13,0.499);
  \draw[cB, thick]
    ([xshift=1.5pt]axis cs:13,0.487) -- ([xshift=10.5pt]axis cs:13,0.487);
 
  \draw[cB, semithick] ([xshift=6pt]axis cs:14,0.453) -- ([xshift=6pt]axis cs:14,0.536);
  \draw[cB, semithick]
    ([xshift=4pt]axis cs:14,0.453) -- ([xshift=8pt]axis cs:14,0.453)
    ([xshift=4pt]axis cs:14,0.536) -- ([xshift=8pt]axis cs:14,0.536);
  \fill[cB, opacity=0.25]
    ([xshift=2.5pt]axis cs:14,0.477) rectangle ([xshift=9.5pt]axis cs:14,0.508);
  \draw[cB, semithick]
    ([xshift=2.5pt]axis cs:14,0.477) rectangle ([xshift=9.5pt]axis cs:14,0.508);
  \draw[cB, thick]
    ([xshift=1.5pt]axis cs:14,0.493) -- ([xshift=10.5pt]axis cs:14,0.493);
 
  \draw[cB, semithick] ([xshift=6pt]axis cs:15,0.447) -- ([xshift=6pt]axis cs:15,0.498);
  \draw[cB, semithick]
    ([xshift=4pt]axis cs:15,0.447) -- ([xshift=8pt]axis cs:15,0.447)
    ([xshift=4pt]axis cs:15,0.498) -- ([xshift=8pt]axis cs:15,0.498);
  \fill[cB, opacity=0.25]
    ([xshift=2.5pt]axis cs:15,0.468) rectangle ([xshift=9.5pt]axis cs:15,0.487);
  \draw[cB, semithick]
    ([xshift=2.5pt]axis cs:15,0.468) rectangle ([xshift=9.5pt]axis cs:15,0.487);
  \draw[cB, thick]
    ([xshift=1.5pt]axis cs:15,0.477) -- ([xshift=10.5pt]axis cs:15,0.477);
 
  \draw[cB, semithick] ([xshift=6pt]axis cs:16,0.402) -- ([xshift=6pt]axis cs:16,0.450);
  \draw[cB, semithick]
    ([xshift=4pt]axis cs:16,0.402) -- ([xshift=8pt]axis cs:16,0.402)
    ([xshift=4pt]axis cs:16,0.450) -- ([xshift=8pt]axis cs:16,0.450);
  \fill[cB, opacity=0.25]
    ([xshift=2.5pt]axis cs:16,0.417) rectangle ([xshift=9.5pt]axis cs:16,0.433);
  \draw[cB, semithick]
    ([xshift=2.5pt]axis cs:16,0.417) rectangle ([xshift=9.5pt]axis cs:16,0.433);
  \draw[cB, thick]
    ([xshift=1.5pt]axis cs:16,0.425) -- ([xshift=10.5pt]axis cs:16,0.425);
 
  \draw[cB, semithick] ([xshift=6pt]axis cs:17,0.383) -- ([xshift=6pt]axis cs:17,0.469);
  \draw[cB, semithick]
    ([xshift=4pt]axis cs:17,0.383) -- ([xshift=8pt]axis cs:17,0.383)
    ([xshift=4pt]axis cs:17,0.469) -- ([xshift=8pt]axis cs:17,0.469);
  \fill[cB, opacity=0.25]
    ([xshift=2.5pt]axis cs:17,0.394) rectangle ([xshift=9.5pt]axis cs:17,0.429);
  \draw[cB, semithick]
    ([xshift=2.5pt]axis cs:17,0.394) rectangle ([xshift=9.5pt]axis cs:17,0.429);
  \draw[cB, thick]
    ([xshift=1.5pt]axis cs:17,0.412) -- ([xshift=10.5pt]axis cs:17,0.412);
 
\addplot[only marks, mark=triangle*, mark size=2.5pt, color=cC, forget plot]
  coordinates {
    (5,0.473)(6,0.542)(7,0.518)(8,0.491)
    (9,0.553)(10,0.474)(11,0.483)
    (12,0.414)(13,0.423)(14,0.482)(15,0.478)(16,0.404)(17,0.376)};
 
\addplot[only marks, mark=star, mark size=3pt, color=cD, semithick, forget plot]
  coordinates {
    (5,0.517)(6,0.556)(7,0.531)(8,0.527)
    (9,0.578)(10,0.613)(11,0.62)
    (12,0.557)(13,0.56)(14,0.581)(15,0.555)(16,0.518)(17,0.536)};

\draw (rel axis cs:0,-0.23) -- (rel axis cs:1,-0.23);

\draw (rel axis cs:0.0402,-0.20) -- (rel axis cs:0.0402,-0.23);
\node[font=\scriptsize] at (rel axis cs:0.0402,-0.30) {57};

\draw (rel axis cs:0.0977,-0.20) -- (rel axis cs:0.0977,-0.23);
\node[font=\scriptsize] at (rel axis cs:0.0977,-0.30) {131};

\draw (rel axis cs:0.1552,-0.20) -- (rel axis cs:0.1552,-0.23);
\node[font=\scriptsize] at (rel axis cs:0.1552,-0.30) {259};

\draw (rel axis cs:0.2126,-0.20) -- (rel axis cs:0.2126,-0.23);
\node[font=\scriptsize] at (rel axis cs:0.2126,-0.30) {511};

\draw (rel axis cs:0.2701,-0.20) -- (rel axis cs:0.2701,-0.23);
\node[font=\scriptsize] at (rel axis cs:0.2701,-0.30) {54};

\draw (rel axis cs:0.3276,-0.20) -- (rel axis cs:0.3276,-0.23);
\node[font=\scriptsize] at (rel axis cs:0.3276,-0.30) {114};

\draw (rel axis cs:0.3851,-0.20) -- (rel axis cs:0.3851,-0.23);
\node[font=\scriptsize] at (rel axis cs:0.3851,-0.30) {212};

\draw (rel axis cs:0.4425,-0.20) -- (rel axis cs:0.4425,-0.23);
\node[font=\scriptsize] at (rel axis cs:0.4425,-0.30) {387};

\draw (rel axis cs:0.5000,-0.20) -- (rel axis cs:0.5000,-0.23);
\node[font=\scriptsize] at (rel axis cs:0.5000,-0.30) {465};

\draw (rel axis cs:0.5575,-0.20) -- (rel axis cs:0.5575,-0.23);
\node[font=\scriptsize] at (rel axis cs:0.5575,-0.30) {55};

\draw (rel axis cs:0.6149,-0.20) -- (rel axis cs:0.6149,-0.23);
\node[font=\scriptsize] at (rel axis cs:0.6149,-0.30) {51};

\draw (rel axis cs:0.6724,-0.20) -- (rel axis cs:0.6724,-0.23);
\node[font=\scriptsize] at (rel axis cs:0.6724,-0.30) {154};

\draw (rel axis cs:0.7299,-0.20) -- (rel axis cs:0.7299,-0.23);
\node[font=\scriptsize] at (rel axis cs:0.7299,-0.30) {155};

\draw (rel axis cs:0.7874,-0.20) -- (rel axis cs:0.7874,-0.23);
\node[font=\scriptsize] at (rel axis cs:0.7874,-0.30) {116};

\draw (rel axis cs:0.8448,-0.20) -- (rel axis cs:0.8448,-0.23);
\node[font=\scriptsize] at (rel axis cs:0.8448,-0.30) {116};

\draw (rel axis cs:0.9023,-0.20) -- (rel axis cs:0.9023,-0.23);
\node[font=\scriptsize] at (rel axis cs:0.9023,-0.30) {23};

\draw (rel axis cs:0.9598,-0.20) -- (rel axis cs:0.9598,-0.23);
\node[font=\scriptsize] at (rel axis cs:0.9598,-0.30) {24};

\legend{
  Llama-3.1-8B-Instruct,
  Qwen2.5-32B-Instruct,
  Llama-3.3-70B-Instruct,
  gpt-5.1,
}

\end{groupplot}
\end{tikzpicture}
\caption{F1-scores for (a)~\textbf{RD Identification} and (b)~\textbf{RD Assessment} across retrieval strategies
         and LLM configurations. Numbers on the ruler refer to the average passage length (in words) associated to the given hyperparameter value. For example: Adaptive strategy with $t=0.3$ produces passages of 465 words on average. Dashed horizontal lines indicate the Long Context (non-retrieval) baseline, in the corrisponding colour.}
\label{fig:results}
\end{figure*}

Given the mixed nature of answer formats (binary, single-answer, and multiple-answer questions) we frame evaluation as a classification problem at the level of individual answer options. Each correct answer option selected by the model is considered a true positive, each option selected by the model but absent from the gold annotation counts as a false positive, and each gold answer option missed by the model counts as a false negative. This per-option evaluation is strict: for a binary question where the model answers \enquote{yes} and the gold answer is \enquote{no}, both a false positive and a false negative are counted. On this basis, we compute Precision, Recall and F1. We report results for the Full Task, as well as separately for the two subtasks: RD Identification and RD Assessment. Results are depicted in Figure \ref{fig:results} and additionally in Appendix \ref{sec:appendix_research_design_analysis}. In what follows, we analyze the results along three dimensions: the impact of retrieval strategy, the role of text embeddings and the variation of performance across Research Designs.

\subsection{Retrieval Strategies}
\label{sec:eval_retrieval_strategies}

When comparing strategies at their best configuration, by selecting the top-performing embedding and hyperparameter combination for each, Adaptive achieves the highest F1 on the Full Task for both \texttt{Llama-8B} (.563) and \texttt{Qwen-32B} (.660), slightly ahead of BM25 (.555 and .645 respectively). However, when we average over embeddings to remove the dependency on a favorable embedding choice, the picture changes: BM25 matches or outperforms all dense strategies across both LLMs and all subtasks, proving to be the most reliable strategy overall (see Table \ref{tab:strategy_comparison}, Appendix \ref{sec:appendix_retrieval_strategies}). 

\begin{table}[!ht]
\centering
\small
\resizebox{\columnwidth}{!}{%
\begin{tabular}{@{}llll@{}}
\toprule
\textbf{LLM} & \textbf{Full Task} & \textbf{RD-I} & \textbf{RD-A} \\
\midrule
Llama-8B  & .779 (p < .001) & .620 (p < .01) & .757 (p < .001) \\
Qwen-32B  & .815 (p < .001) & .703 (p < .01) & .810 (p < .001) \\
Llama-70B & .727 (p < .01)  & .532 (p < .05) & .655 (p < .01) \\
gpt-5.1   & .292 (p = n.s.) & .534 (p < .05) & .299 (p = n.s.) \\
\bottomrule
\end{tabular}
}
\caption{Pearson correlation ($r$) between average chunk length and F1 score across the 17 hyperparameter configurations. For \texttt{Llama-8B} and \texttt{Qwen-32B}, F1 scores of dense strategies are averaged over 6 embedding models.}
\label{tab:chunk_length_correlation}
\end{table}

Within each strategy, then, performance is maximized around certain hyperparameters, i.e. 512-1024 for fixed-size strategies (BM25 and Dense), $t = 0.3$ for Adaptive and $st = 0.3$ for Propositional. Sensitivity to hyperparameters is much higher in variable-size strategies, in particular higher values of $t$ for Adaptive and $st$ for Propositional (which in both cases yield shorter passages) cause a significant drop in performance, and this holds in both RD-I and RD-A. To test whether passage length, rather than other strategy related properties, drive these differences, we first computed the average passage length (Figure \ref{fig:results}, ruler) produced by each hyperparameter value for all the strategies across the whole dataset and then correlated them with the performance. This reveals that the overall performance, at least for the three smallest LLMs, has a high linear correlation with average passage length: 0.779 Pearson \textit{r} for \texttt{Llama-8B}, 0.815 for \texttt{Qwen-32B} and 0.727 for \texttt{Llama-70B} respectively. Correlation values are significantly higher for Assessment than for Identification (see Table \ref{tab:chunk_length_correlation}). Different evidence holds for \texttt{gpt-5.1}, whose performance does not correlate significantly with passage length. Crucially, this provides a very clear indication for our task: between 52\% and 66\% of the task variance (R-squared) is explained by chunk length, showing that context is relevant to enable correct answering patterns in a RAG-baseline pipeline. Also, the lack of correlation for \texttt{gpt-5.1} suggests that very strong LLMs can compensate for the lack of context, possibly with a better use of conversational history or via parametric knowledge. Since the observed correlation between passage length and performance could in principle be confounded with other strategy-related properties, we conducted a further analysis by comparing different strategies \textit{at matched passage lengths}. The underlying idea is that, if a given strategy holds a systematic advantage over another, this should be reflected in consistent performance differences when comparing configurations that yield similar average passage lengths.
We compared 5 pairs of configurations (reported in Table \ref{tab:configs_matched_length}) with \texttt{Llama-8B} and \texttt{Qwen-32B}, for which we have results across all 6 embedding models, allowing us to average out the effect of embedding choice. Across 30 comparisons (5 pairs with 2 LLMs and 3 tasks), the average performance difference between strategies is 1.0 F1-points (with the worst case lying at 3.1 points) and no strategy shows a consistent advantage over the others. Overall, this indicates that, while differences in strategies might exist, they probably lie at other levels (lexical, boundaries), and that passage length is the strongest driver for performance across strategies.

\begin{table}[!ht]
\centering
\small
\resizebox{\columnwidth}{!}{%
\begin{tabular}{@{}clccll@{}}
\toprule
& \textbf{Config A} & len & vs. & \textbf{Config B} & len\\
\midrule
(a) & Dense 128 & 54 & & Adaptive $t$=0.5& 55\\
(b) & BM25 256 & 131 & & Propositional $st$=0.3, $ws$=2& 154\\
(c) & Dense 256 & 114 & & Propositional $st$=0.5 & 116\\
(d) & Dense 1024 & 387 & & Adaptive $t$=0.3& 465\\
(e) & BM25 1024 & 511 & & Adaptive $t$=0.3 & 465\\
\bottomrule
\end{tabular}
}
\caption{Values for the 5 configuration pairs used for matched length comparison. The \textit{len} columns refer to the average passage length generated by the given configuration, measured in words (split on whitespace).}
\label{tab:configs_matched_length}
\end{table}

\subsection{Text Embeddings}
\label{sec:text_embeddings}
Using the F1 scores from single experimental rounds with \texttt{Llama-8B} and \texttt{Qwen-32B}, we aggregated the results of different embedding models by subtask, retrieval strategy and across strategy parameters. Based on simple win count (see Table \ref{tab:embedding_wins}, Appendix \ref{sec:appendix_text_embeddings}), \texttt{bge} leads the ranking, being the best model in 41.5\% of cases, followed by \texttt{SFR} (20.7\%) and \texttt{mxbai} (13.4\%). This trend holds for Dense and Adaptive strategies, while for Propositional \texttt{SFR} clearly dominates. As for subtasks, \texttt{bge} is noticeably the best in the RD-I task and in the Full Task, while in RD-A it ties with \texttt{SFR}. \texttt{bge} also shows the top mean F1 across the three subtasks and the second lowest standard deviation across hyperparameters, after \texttt{SFR}. Overall, \texttt{bge} offers the best trade-off between performance and stability among the models tested. Looking at the broader picture, we observe that the gap between the best and the worst embedding drops with larger LLMs, passing from 2.5 to 2.1 F1 points in the RD-I task and from 4.7 to 3.3 points in the more difficult RD-A task. A similar but less pronounced trend is followed by the standard deviation, which decreases from 0.010 to 0.008 in RD-I and from 0.017 to 0.014 in RD-A (see Table \ref{tab:embedding_impact}). Both observations, combined, indicate that the choice of embedding is less relevant with stronger models, but also that it becomes more important with the growing difficulty of the task.

\subsection{Research Design Analysis}

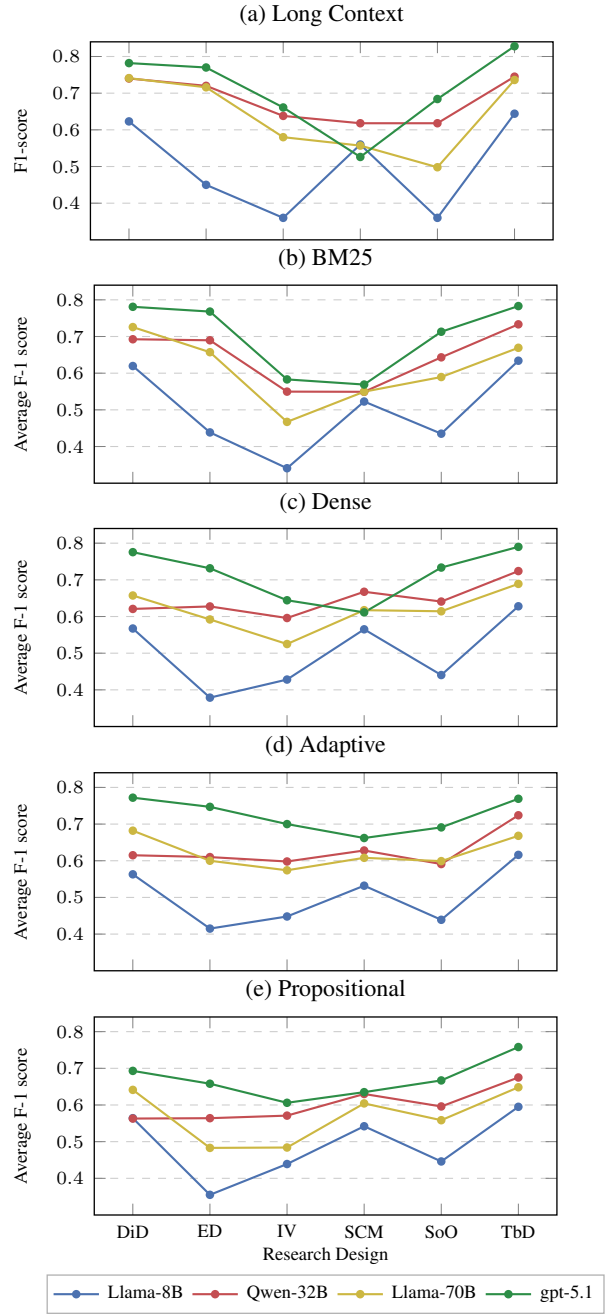
\begin{figure}[!h]
\centering

\pgfplotsset{
    shared style/.style={
        width=\linewidth,
        height=4.2cm,
        xmin=0.5, xmax=6.5,
        ymin=0.3, ymax=0.84,
        xtick={1,2,3,4,5,6},
        ytick={0.4,0.5,0.6,0.7,0.8},
        ylabel={Average F-1 score},
        ylabel style={font=\scriptsize, yshift=2pt},
        yticklabel style={font=\scriptsize},
        ymajorgrids=true,
        grid style={dashed, gray!40},
        title style={font=\small, yshift=-4pt},
        tick label style={font=\scriptsize},
        every axis/.append style={line width=0.5pt},
    }
}

\begin{subfigure}{\linewidth}
    \begin{tikzpicture}
        \begin{axis}[
            shared style,
            title={(a) Long Context},
            xticklabels=\empty,
            ylabel={F1-score}
        ]
        \addplot[color=cA, solid, mark=*, mark size=1.2pt, line width=0.8pt,
            mark options={fill=cA, solid}]
            coordinates {(1,0.623)(2,0.45)(3,0.36)(4,0.56)(5,0.36)(6,0.644)};
        \addplot[color=cB, solid, mark=*, mark size=1.2pt, line width=0.8pt,
            mark options={fill=cB, solid}]
            coordinates {(1,0.74)(2,0.72)(3,0.638)(4,0.618)(5,0.618)(6,0.745)};
        \addplot[color=cC, solid, mark=*, mark size=1.2pt, line width=0.8pt,
            mark options={fill=cC, solid}]
            coordinates {(1,0.741)(2,0.716)(3,0.58)(4,0.557)(5,0.498)(6,0.736)};
        \addplot[color=cD, solid, mark=*, mark size=1.2pt, line width=0.8pt,
            mark options={fill=cD, solid}]
            coordinates {(1,0.782)(2,0.77)(3,0.661)(4,0.526)(5,0.684)(6,0.828)};
    \end{axis}
    \end{tikzpicture}
    \label{fig:chart_a}
\end{subfigure}

\vspace{-0.8em}

\begin{subfigure}{\linewidth}
    \begin{tikzpicture}
        \begin{axis}[
            shared style,
            title={(b) BM25},
            xticklabels=\empty,
        ]
        \addplot[color=cA, solid, mark=*, mark size=1.2pt, line width=0.8pt,
            mark options={fill=cA, solid}]
            coordinates {(1,0.6195)(2,0.4385)(3,0.3408)(4,0.5227)(5,0.435)(6,0.634)};
        \addplot[color=cB, solid, mark=*, mark size=1.2pt, line width=0.8pt,
            mark options={fill=cB, solid}]
            coordinates {(1,0.6925)(2,0.6897)(3,0.5497)(4,0.5492)(5,0.6432)(6,0.7332)};
        \addplot[color=cC, solid, mark=*, mark size=1.2pt, line width=0.8pt,
            mark options={fill=cC, solid}]
            coordinates {(1,0.7257)(2,0.6572)(3,0.4672)(4,0.5495)(5,0.5895)(6,0.6692)};
        \addplot[color=cD, solid, mark=*, mark size=1.2pt, line width=0.8pt,
            mark options={fill=cD, solid}]
            coordinates {(1,0.781)(2,0.768)(3,0.583)(4,0.569)(5,0.713)(6,0.783)};
    \end{axis}
    \end{tikzpicture}
    \label{fig:chart_b}
\end{subfigure}

\vspace{-0.8em}

\begin{subfigure}{\linewidth}
    \begin{tikzpicture}
        \begin{axis}[
            shared style,
            title={(c) Dense},
            xticklabels=\empty,
        ]
        \addplot[color=cA, solid, mark=*, mark size=1.2pt, line width=0.8pt,
            mark options={fill=cA, solid}]
            coordinates {(1,0.5673)(2,0.3788)(3,0.4281)(4,0.5651)(5,0.4404)(6,0.6279)};
        \addplot[color=cB, solid, mark=*, mark size=1.2pt, line width=0.8pt,
            mark options={fill=cB, solid}]
            coordinates {(1,0.6207)(2,0.6275)(3,0.5959)(4,0.6675)(5,0.6408)(6,0.7239)};
        \addplot[color=cC, solid, mark=*, mark size=1.2pt, line width=0.8pt,
            mark options={fill=cC, solid}]
            coordinates {(1,0.6573)(2,0.592)(3,0.525)(4,0.6173)(5,0.6143)(6,0.689)};
        \addplot[color=cD, solid, mark=*, mark size=1.2pt, line width=0.8pt,
            mark options={fill=cD, solid}]
            coordinates {(1,0.7755)(2,0.7315)(3,0.6443)(4,0.6113)(5,0.7335)(6,0.790)};
    \end{axis}
    \end{tikzpicture}
    \label{fig:chart_c}
\end{subfigure}

\vspace{-0.8em}

\begin{subfigure}{\linewidth}
    \begin{tikzpicture}
        \begin{axis}[
            shared style,
            title={(d) Adaptive},
            xticklabels=\empty,
        ]
        \addplot[color=cA, solid, mark=*, mark size=1.2pt, line width=0.8pt,
            mark options={fill=cA, solid}]
            coordinates {(1,0.563)(2,0.415)(3,0.448)(4,0.532)(5,0.439)(6,0.616)};
        \addplot[color=cB, solid, mark=*, mark size=1.2pt, line width=0.8pt,
            mark options={fill=cB, solid}]
            coordinates {(1,0.615)(2,0.610)(3,0.598)(4,0.628)(5,0.591)(6,0.724)};
        \addplot[color=cC, solid, mark=*, mark size=1.2pt, line width=0.8pt,
            mark options={fill=cC, solid}]
            coordinates {(1,0.682)(2,0.600)(3,0.574)(4,0.608)(5,0.599)(6,0.668)};
        \addplot[color=cD, solid, mark=*, mark size=1.2pt, line width=0.8pt,
            mark options={fill=cD, solid}]
            coordinates {(1,0.772)(2,0.747)(3,0.700)(4,0.662)(5,0.691)(6,0.769)};
    \end{axis}
    \end{tikzpicture}
    \label{fig:chart_d}
\end{subfigure}

\vspace{-0.8em}

\begin{subfigure}{\linewidth}
    \begin{tikzpicture}
        \begin{axis}[
            shared style,
            title={(e) Propositional},
            xticklabels={DiD, ED, IV, SCM, SoO, TbD},
            xticklabel style={font=\scriptsize},
            xlabel={Research Design},
            xlabel style={font=\scriptsize, yshift=4pt},
            legend entries={Llama-8B, Qwen-32B, Llama-70B, gpt-5.1},
            legend style={
                at={(0.5, -0.3)},
                anchor=north,
                font=\scriptsize,
                legend columns=4,
                column sep=1pt,
                inner sep=2pt,
                draw=gray!60,
            },
        ]
        \addplot[color=cA, solid, mark=*, mark size=1.2pt, line width=0.8pt,
            mark options={fill=cA, solid}]
            coordinates {(1,0.564)(2,0.355)(3,0.439)(4,0.542)(5,0.446)(6,0.595)};
        \addplot[color=cB, solid, mark=*, mark size=1.2pt, line width=0.8pt,
            mark options={fill=cB, solid}]
            coordinates {(1,0.563)(2,0.564)(3,0.571)(4,0.630)(5,0.596)(6,0.675)};
        \addplot[color=cC, solid, mark=*, mark size=1.2pt, line width=0.8pt,
            mark options={fill=cC, solid}]
            coordinates {(1,0.6412)(2,0.483)(3,0.484)(4,0.6043)(5,0.5585)(6,0.6483)};
        \addplot[color=cD, solid, mark=*, mark size=1.2pt, line width=0.8pt,
            mark options={fill=cD, solid}]
            coordinates {(1,0.693)(2,0.658)(3,0.606)(4,0.635)(5,0.667)(6,0.758)};
    \end{axis}
    \end{tikzpicture}
    \label{fig:chart_e}
\end{subfigure}

\caption{Performance of four LLMs, by RD. Except for the Long Context baseline, results for
\texttt{Llama-8B} and \texttt{Qwen-32B} refer to the average performance of
6 embedding models for the given RD. \texttt{Llama-70B} and \texttt{gpt-5.1}
used only \texttt{bge-large-en-v1.5}.}
\label{fig:all_charts}

\end{figure}

Based on their specific methodological architecture, Research Designs differ in many respects: in their degree of formalization, the type of technical language used, the way argumentation is constructed and the density of relevant information throughout the paper. We therefore analyzed the performance per-RD, by computing performance for each LLM (across strategies, Figure \ref{fig:all_charts}) and a ranking for each strategy (see Table \ref{tab:rd_ranking}).   
The results provide a nuanced picture. At the two opposite ends of the spectrum, we observe that TbD are consistently the easiest to identify and assess by models, followed by DiD, while IV are quite consistently the most difficult. SCM and ED, in turn, are more sensitive to the specific LLM. SoO, meanwhile, shows a more consistent ranking across all four models. When results are grouped by strategy, BM25 appears to show the highest variability across RDs, possibly because its lexical matching is more sensitive to how explicitly each design's methodology is described.

\begin{table}[h]
\centering
\small
\begin{tabular}{@{}lcccccc@{}}
\toprule 
\textbf{RD} & \texttt{L-8B} & \texttt{Q-32B} & \texttt{L-70B} & \texttt{gpt-5.1} & \textbf{MR} & \textbf{Std} \\ 
\midrule 
TbD & 1 & 1 & 3 & 1 & 1.5 & 0.9 \\ 
DiD & 2 & 2 & 1 & 2 & 1.8 & 0.4 \\ 
SCM & 3 & 4 & 2 & 6 & 3.8 & 1.5 \\ 
SoO & 4 & 5 & 4 & 4 & 4.2 & 0.4 \\ 
ED & 6 & 3 & 5 & 3 & 4.2 & 1.3 \\ 
IV & 5 & 6 & 6 & 5 & 5.5 & 0.5 \\ 
\bottomrule 
\end{tabular} 
\caption[Per-RD difficulty ranking across LLMs]{Difficulty ranking of the six RDs (rank~1 = easiest) for each LLM. For each (LLM, RD) pair, F1 is averaged within each retrieval strategy and then across the four strategies (equal weight per strategy); RDs are then ranked within each LLM. \textbf{Mean Rank} and \textbf{Std} are computed across the four LLM ranks.} 
\label{tab:rd_ranking}
\end{table}

We further correlated each of the 4 sets of F1-averages per Research Design with the human agreement, in order to test the hypothesis that RDs that are difficult for humans to annotate would also be harder for models to predict correctly. As showed in Table \ref{tab:agreement-performance}, \textit{we observed no statistically significant correlation}. Indeed, several cases exhibited an inverse pattern, with RDs characterized by lower human agreement being comparatively easier for models to predict (for example DiD), and vice versa (e.g. ED). The only RD where human and machine difficulty align is IV, which proves challenging for both. Although surprising, this suggests that the RDs on which humans disagree the most are not the same on which automated system struggle the most.
In other words, human difficulty, i.e. the subjectivity inherent to expert judgment, and machine difficulty, that of correctly retrieving the right passages from the text, appear to be two largely independent dimensions in expert tasks like ARDTrA. This finding is in line with recent debate in IR and NLP \citep{amiraz-etal-2025-distracting, trappolini-etal-2026-redefining}, who show that at the \textit{passage} level, human relevance judgments and actual machine utility in RAG systems are fundamentally misaligned. Our results suggest that a similar misalignment exists at the \textit{task} level. Understanding how this misalignment operates remains an open challenge, with direct implications for the design and evaluation of RAG systems in expert domains.

\begin{table}[!h]
\centering
\small
\setlength{\tabcolsep}{3.5pt}
\begin{tabular}{lcccc}
\toprule
 & \textbf{Llama-8B} & \textbf{Qwen-32B} & \textbf{Llama-70B} & \textbf{gpt-5.1} \\
\midrule
BM25      & .27/.49 & .06/-.09 & .14/.09 & -.04/-.09 \\
Dense     & .21/.09 & .49/.43   & .21/.26 & -.20/-.31 \\
Adaptive  & .20/.14 & .44/.71   & .04/.43 & .00/-.09 \\
TopicNode & .05/.09 & .40/.20   & .07/.09 & .12/-.09 \\
\bottomrule
\end{tabular}
\caption{Pearson $r$ / Spearman $\rho$ correlations between
Krippendorff's $\alpha$ and model performance across the six Research Designs. None of the correlations is statistically significant ($p > .05$).}
\label{tab:agreement-performance}
\end{table}

\subsection{Long Context vs. RAG}
Long-Context document prompting proves a strong baseline: across both subtasks and all four LLMs, most retrieval configurations fail to outperform it (dashed lines in Figure \ref{fig:results}). On RD-I, the baseline is only locally surpassed, and mainly by the smaller models, suggesting that when the task is comparatively easy, retrieval offers little advantage over simply reading the whole document. On RD-A, the picture is similar but the exceptions are more systematic: only BM25 and the Adaptive strategy manage to beat the LC baseline at specific configurations. These results indicate that retrieval is not, in itself, necessary to achieve competitive performance on this task; rather, its value lies in focusing the model on a small set of passages that can be stored and inspected, providing the interpretability that a black-box LC pipeline cannot.

\section{Conclusions and Future Work}
\label{sec:conclusions}
In this work, we introduced the task of Automatic Research Design Tracking and Assessment (ARDTrA), meant to help social scientists and policy advisors quickly and reliably estimate the robustness of scientific papers based on an extensive analysis of their application of Research Designs. This contribution bridges a gap in existing NLP-CSS research, in particular by providing an expert-designed analytical framework and an annotated dataset of 140 papers from the social sciences. We evaluated four different retrieval strategies, based on fixed and variable chunking, with four LLMs and six embedding models, as well as a Long Context setting.
Based on extensive experimental evidence, our findings show that: (a) low-cost, sparse BM25-based retrieval matches and often outperforms dense embedding-based strategies; (b) passage length, rather than retrieval strategy, explains between 52\% and 66\% of the performance variance across configurations, and at matched passage lengths different strategies perform comparably, though this effect diminishes with stronger LLMs such as \texttt{gpt-5.1}, whose performance is largely independent of passage length; (c) for embedding-based strategies, the impact of embedding model choice is smaller than that of retrieval strategy or hyperparameter configuration, with smaller encoder-based models like \texttt{bge-large-en-v1.5} proving the most robust across settings; moreover, (d) Long-Context document prompting proves to be a strong baseline when compared to RAG. Finally, (e) per-RD analysis reveals that human and machine difficulty are driven by different factors: the Research Designs on which experts disagree most are not the same ones on which the automated system struggles, suggesting that subjective judgment and information retrievability represent two independent dimensions of task difficulty.
Findings (b) and (e), in particular, open up promising research directions that we aim to address in future work. First, while passage length accounts for a substantial share of the variance, a significant portion remains unexplained; we plan to explore whether lexical factors (how methodological choices are expressed linguistically) and structural ones (where in the document the relevant information appears) contribute to the residual variance. Second, our finding that human and machine difficulty diverge raises the question of whether experts and retrieval-based systems attend to fundamentally different textual cues when performing methodological assessment, and whether this divergence generalizes to other expert analytical tasks.

\section{Limitations}
\label{sec:limitations}

The main limitation of this work is the size of the dataset. This is due to the depth of the annotation schema and the high effort required from annotators, who must closely read papers of up to 50 pages and answer subtle methodological questions. With the current size of the dataset, therefore, per-RD estimates should be interpreted with caution. We tried to counterbalance the limited scale by carefully selecting papers to be representative of each RD and to cover a range of application domains. Additionally, the dataset focuses on papers in English, which limits the generalizability of our findings to other languages.

\section{Use of AI Assistants}

We used AI Assistants to support the revision of the manuscript and the creation of complex plots/tables. All  scientific content, experimental design, data analysis, and interpretation of results are the sole responsibility of the authors.

\section{Ethics Statement}
This work analyzes published scientific papers and does not involve human subjects, personal data, or sensitive content. All annotators are co-authors and participated as part of their regular institutional duties. The dataset release (see Supplemental Materials) contains only bibliographic metadata as most of the papers are copyrighted. To the best of our knowledge, the work described in this paper does not raise any ethical concerns.

\bibliography{biblio}

\appendix
\clearpage
\section{Retrieval Strategies}
\label{sec:appendix_retrieval_strategies}

In this Section we report the results of the analysis conducted on Retrieval Strategies (Section \ref{sec:eval_retrieval_strategies}).

\begin{table}[!ht]
\centering
\small
\resizebox{\columnwidth}{!}{%
\begin{tabular}{@{}llcccc@{}}
\toprule
\textbf{LLM} & \textbf{Task} & \textbf{BM25} & \textbf{DENSE} & \textbf{ADAPT} & \textbf{PROP} \\
\midrule
\multicolumn{6}{@{}l}{\textit{Best overall score}} \\
\midrule
\multirow{3}{*}{Ll-8B}
& Full Task & .555 & .541 (\texttt{b}) & \cellcolor{blue!15}\textbf{.563} (Q) & .543 (S) \\
& RD-I      & .712 & .727 (b) & .729 (Q) & \cellcolor{blue!15}\textbf{.742} (e) \\
& RD-A      & .443 & .442 (b) & \cellcolor{blue!15}\textbf{.447} (Q) & .409 (S) \\
\midrule
\multirow{3}{*}{Qw-32B}
& Full Task & .645 & .641 (b) & \cellcolor{blue!15}\textbf{.660} (b) & .629 (m) \\
& RD-I      & .773 & .764 (b) & .767 (m) & \cellcolor{blue!15}\textbf{.774} (S) \\
& RD-A      & .557 & .562 (Q) & \cellcolor{blue!15}\textbf{.590} (b) & .536 (m) \\
\midrule
\multicolumn{6}{@{}l}{\textit{Best score embedding average}} \\
\midrule
\multirow{3}{*}{Ll-8B}
& Full Task & \cellcolor{blue!15}\textbf{.555} & .526 & .539 & .513 \\
& RD-I      & .712 & \cellcolor{blue!15}\textbf{.715} & .715 & .711 \\
& RD-A      & \cellcolor{blue!15}\textbf{.443} & .393 & .421 & .375 \\
\midrule
\multirow{3}{*}{Qw-32B}
& Full Task & \cellcolor{blue!15}\textbf{.645} & .627 & .630 & .607 \\
& RD-I      & \cellcolor{blue!15}\textbf{.773} & .753 & .715 & .759 \\
& RD-A      & \cellcolor{blue!15}\textbf{.557} & .542 & .544 & .504 \\
\midrule
\multicolumn{6}{@{}l}{\textit{Single embedding (\texttt{bge-large-en-v1.5})}} \\
\midrule
\multirow{3}{*}{Ll-70B}
& Full Task & \cellcolor{blue!15}\textbf{.640} & .626 & .634 & .589 \\
& RD-I      & .743 & .749 & \cellcolor{blue!15}\textbf{.754} & .742 \\
& RD-A      & \cellcolor{blue!15}\textbf{.573} & .542 & .553 & .482 \\
\midrule
\multirow{3}{*}{gpt-5.1}
& Full Task & \cellcolor{blue!15}\textcolor{green!60!black}{\textbf{.694}} & .658 & .686 & .664 \\
& RD-I      & .811 & \cellcolor{blue!15}\textcolor{green!60!black}{\textbf{.814}} & .803 & .813 \\
& RD-A      & .618 & .556 & \cellcolor{blue!15}\textcolor{green!60!black}{\textbf{.620}} & .581 \\
\bottomrule
\end{tabular}
} 
\caption[Strategy comparison (F1-scores)]{Strategy comparison (F1-scores). \colorbox{blue!15}{Highlight} indicates the best strategy per row. \textit{Best embedding}: best hyperparameter configuration with the best-performing embedding model. \textit{Embedding avg.}: best hyperparameter configuration, averaged over all 6 embedding models. Llama-70B and gpt-5.1 use \texttt{bge-large-en-v1.5} only.}
\label{tab:strategy_comparison}
\end{table}

\section{Text Embeddings}
\label{sec:appendix_text_embeddings}
This section refers to the analysis on the impact of text embeddings conducted in Section \ref{sec:text_embeddings} of the paper. Values in Table \ref{tab:embedding_impact} have been computed by first averaging all F1-scores by embedding model across strategies and then by calculating the max-min range and standard deviation. Values in Table \ref{tab:embedding_wins} have been computed by counting the times each embedding model provides the best F1 score for a given hyperparameter. As for Table \ref{tab:embedding_comparison}, Mean F1 values have been computed by averaging the given model performance across all hyperparameters and two LLMs. Standard deviation and Range, instead have been computed by first calculating Std value for each LLM in RD-I and RD-A tasks for the given embedding model, thus obtaining 4 values for each embedding model. Final Std in the Table is the average of these values, while Range is their max-min difference.

\label{sec:appendix_text_embeddings}
\begin{table}[!ht]
\centering
\small
\begin{tabular}{@{}llcc@{}}
\toprule
\textbf{LLM} & \textbf{Task} & \textbf{Range} & \textbf{Std} \\
\midrule

\multirow{3}{*}{\texttt{Llama-8B}}
& Full Task & .028 & .011 \\
& RD-I      & .025 & .010 \\
& RD-A      & .047 & .017 \\

\midrule

\multirow{3}{*}{\texttt{Qwen-32B}}
& Full Task & .024 & .010 \\
& RD-I      & .021 & .008 \\
& RD-A      & .033 & .014 \\

\bottomrule
\end{tabular}
\caption{Embedding impact: range and standard deviation of F1 scores across the six embedding models, by task and LLM.}
\label{tab:embedding_impact}
\end{table}

\begin{table}[!ht]
\centering
\small
\resizebox{\columnwidth}{!}{%
\begin{tabular}{@{}lcccccc@{}}
\toprule
& \textbf{bge} & \textbf{SFR} & \textbf{mxbai} & \textbf{Qwen3} & \textbf{e5} & \textbf{Cohere} \\
\midrule
Overall (78)   & \textbf{34} & 17 & 11 & 10 & 8 & 2 \\
\midrule
Dense (24)     & \textbf{14} & 2  & 3  & 6  & 1 & 0 \\
Adaptive (18)  & \textbf{12} & 1  & 4  & 3  & 0 & 0 \\
Propositional (36) & 8            & \textbf{14} & 4 & 1 & 7 & 2 \\
\midrule
Full Task (26) & \textbf{13} & 5 & 4 & 3 & 1 & 1 \\
RD-I (26)      & \textbf{13} & 4 & 2 & 1 & 7 & 0 \\
RD-A (26)      & 8 & \textbf{8} & 5 & 6 & 0 & 1 \\
\bottomrule
\end{tabular}
}
\caption[Number of times each embedding model achieves the highest F1 score]{Number of times each embedding model achieves the highest F1 score, out of 78 combinations (13 hyperparameter configurations $\times$ 2 LLMs $\times$ 3 tasks). This information is provided in Tables \ref{tab:full_results_appendix}, \ref{tab:rdi_results_appendix} and \ref{tab:rda_results_appendix} in Appendix \ref{sec:appendix_research_design_analysis}. Since there are a few ties, some rows don't sum up to the same number.}
\label{tab:embedding_wins}
\end{table}

\begin{table}[!ht]
\centering
\small
\resizebox{\columnwidth}{!}{%
\begin{tabular}{@{}lccccccc@{}}
\toprule
& \textbf{Size} & \multicolumn{3}{c}{\textbf{Mean F1}} & \multicolumn{2}{c}{\textbf{HP Sensitivity}} \\
\cmidrule(lr){3-5} \cmidrule(lr){6-7}
\textbf{Model} & & \textbf{Full} & \textbf{RD-I} & \textbf{RD-A} & \textbf{Std} & \textbf{Range} \\
\midrule
bge     & 335M  & \textbf{.560} & \textbf{.729} & \textbf{.442} & .029 & .110 \\
SFR     & 7B    & .551 & .719 & .437 & \textbf{.026} & .092 \\
mxbai   & 335M  & .550 & .720 & .433 & .031 & .117 \\
Qwen3   & 8B    & .541 & .707 & .434 & .029 & .097 \\
e5      & 335M  & .536 & .724 & .404 & .029 & \textbf{.089} \\
Cohere  & prop. & .536 & .709 & .416 & .028 & \textbf{.089} \\
\bottomrule
\end{tabular}
}
\caption{Embedding model comparison: mean F1 and sensitivity to hyperparameter configuration. Mean is computed across all configurations (13 hyperparameters $\times$ 2 LLMs). Std.\ and range measure variability across the 13 hyperparameter configurations, averaged over 2 LLMs $\times$ 2 tasks.}
\label{tab:embedding_comparison}
\end{table}

\section{Research Design Analysis}
\label{sec:appendix_research_design_analysis}

This Section reports the raw experimental results before any aggregation (F1-scores). Each score in Tables \ref{tab:full_results_appendix}, \ref{tab:rdi_results_appendix} and \ref{tab:rda_results_appendix} refers to a complete run of the system on the whole dataset. Table \ref{tab:full_results_appendix} refers to the Full Task, Table \ref{tab:rdi_results_appendix} to RD Identification and Table \ref{tab:rda_results_appendix} to RD Assessment.

\begin{table*}[t!]

\centering

\scriptsize

\setlength{\tabcolsep}{4pt}

\begin{tabular}{@{}cccccccl cccc ccc cccccc@{}}

\toprule

& \textbf{LC}
& \multicolumn{4}{c}{\textbf{BM25}}

&
& \multicolumn{4}{c}{\textbf{Dense}}
& \multicolumn{3}{c}{\textbf{Adaptive}}
& \multicolumn{6}{c}{\textbf{Topicnode}} \\

\cmidrule(lr){2-2}
\cmidrule(lr){3-6}
\cmidrule(lr){8-11}
\cmidrule(lr){12-14}
\cmidrule(lr){15-20}

& 
& \multicolumn{4}{c}{chunk size}
&
& \multicolumn{4}{c}{chunk size}
& \multicolumn{3}{c}{t}
& \multicolumn{2}{c}{st=0.3}
& \multicolumn{2}{c}{st=0.5}
& \multicolumn{2}{c}{st=0.8} \\

\cmidrule(lr){3-6}
\cmidrule(lr){8-11}
\cmidrule(lr){12-14}
\cmidrule(lr){15-16}
\cmidrule(lr){17-18}
\cmidrule(lr){19-20}

\textbf{LLM}

& 
& 128 & 256 & 512 & 1024

& \textbf{Emb.}
& 128 & 256 & 512 & 1024

& 0.3 & 0.5 & 0.7

& {\scriptsize ws=2} & {\scriptsize ws=5}
& {\scriptsize ws=2} & {\scriptsize ws=5}
& {\scriptsize ws=2} & {\scriptsize ws=5} \\

\midrule

\multirow{6}{*}{\rotatebox[origin=c]{90}{Llama-8B}}*

& \fbox{.545}
& .483 & .524 & \textbf{.555} & .533
& mxbai
& .503 & .495 & .516 & .525
& .538 & .489 & \cellcolor{blue!15}{.501}
& .517 & \cellcolor{blue!15}{.526}
& .505 & .511
& .455 & .440 \\*

& 
&      &      &      &      
& bge
& .494 & \cellcolor{blue!15}.541 & \cellcolor{blue!15}.540 & \cellcolor{blue!15}.539
& .542 & \cellcolor{blue!15}{.504} & .494
& .511 & .511
& .537 & .532
& \cellcolor{blue!15}{.497} & .468 \\*

& 
&      &      &      &      
& e5
& .444 & .494 & .503 & .512
& .535 & .448 & .452
& \cellcolor{blue!15}{.529} & .466
& .495 & .538
& .459 & .470 \\*

& 
&      &      &      &      
& SFR
& .512 & .520 & .534 & .532
& .543 & .496 & .484
& .524 & .519
& \cellcolor{blue!15}.543 & \cellcolor{blue!15}.542
& .441 & .471 \\*

& 
&      &      &      &      
& Qwen3
& \cellcolor{blue!15}{.513} & .511 & .530 & \cellcolor{blue!15}.539
& \cellcolor{blue!15}\textbf{\uline{.563}} & .486 & .481
& .513 & .516
& .487 & .458
& .452 & .457 \\*

& 
&      &      &      &      
& Cohere
& .496 & .508 & .533 & .511
& .515 & .471 & .479
& .469 & .509
& .488 & .496
& .463 & \cellcolor{blue!15}{.475} \\*

\midrule

\multirow{6}{*}{\rotatebox[origin=c]{90}{Qwen-32B}}

& \fbox{\uline{.674}}
& .598 & .626 & .638 & .645
& mxbai
& .572 & .622 & \cellcolor{blue!15}.626 & .633
& .653 & .582 & .589
& .612 & .603
& \cellcolor{blue!15}{.629} & .597
& .553 & .516 \\

& 
&      &      &      &      
& bge
& \cellcolor{blue!15}{.593} & \cellcolor{blue!15}{.627} & .622 & \cellcolor{blue!15}{.641}
& \cellcolor{blue!15}{.660} & \cellcolor{blue!15}{.592} & \cellcolor{blue!15}{.595}
& .613 & .603
& .602 & \cellcolor{blue!15}{.601}
& \cellcolor{blue!15}{.567} & .522 \\

& 
&      &      &      &      
& e5
& .565 & .590 & .625 & .619
& .610 & .555 & .550
& .598 & .594
& .590 & .596
& .552 & \cellcolor{blue!15}{.550} \\

& 
&      &      &      &      
& SFR
& .561 & .590 & .590 & .617
& .604 & .586 & .560
& \cellcolor{blue!15}{.617} & \cellcolor{blue!15}{.606}
& .625 & .595
& .552 & .573 \\

& 
&      &      &      &      
& Qwen3
& .592 & .565 & .614 & .635
& .634 & .560 & .563
& .602 & .589
& .556 & .566
& .535 & .540 \\

& 
&      &      &      &      
& Cohere
& .561 & .617 & .622 & .618
& .619 & .565 & .552
& .600 & .570
& .580 & .562
& .535 & .527 \\

\midrule

{\rotatebox[origin=c]{90}{70B}}

& \fbox{.636}
& .529 & .611 & .621 & \textbf{\uline{.640}}
& bge
& .574 & .626 & .610 & .593
& .634 & .571 & .576
& .575 & .584
& .589 & .587
& .530 & .513 \\

\midrule

{\rotatebox[origin=c]{90}{5.1}}

& \fbox{.691}
& .643 & .675 & \textcolor{green!60!black}{\textbf{\uline{.694}}} & .672
& bge
& .618 & .658 & .635 & .642
& .670 & .683 & .686
& .654 & .661
& .664 & .655
& .620 & .634 \\

\midrule

\multicolumn{1}{@{}l}{Avg.\ p.\ len}

& -- 
& 57 & 131 & 269 & 511

&
& 54 & 114 & 212 & 387

& 465 & 55 & 51

& 154 & 155 & 116 & 116 & 23 & 24 \\

\bottomrule

\end{tabular}

\caption[Full Task F1 scores across strategies and models]{\textbf{Full Task} F1 scores across retrieval strategies, hyperparameters, embedding models and LLMs. Highlight marks \colorbox{blue!15}{the model scoring the best result for the given hyperparameter}, in boldface the \textbf{configurations that outperform the Long Context (LC) \fbox{baseline}}. Underlined, the best result for the current LLM. In green, the \textcolor{green!60!black}{best overall result for the task}. The last row reports the average passage length generated by the hyperparameter (in number of words).}

\label{tab:full_results_appendix}

\end{table*}

\begin{table*}[t!]

\centering

\scriptsize

\setlength{\tabcolsep}{4pt}

\begin{tabular}{@{}lccccc l cccc ccc cccccc@{}}

\toprule

& \multicolumn{1}{c}{\textbf{LC}}
& \multicolumn{4}{c}{\textbf{BM25}}

&
& \multicolumn{4}{c}{\textbf{Dense}}

& \multicolumn{3}{c}{\textbf{Adaptive}}

& \multicolumn{6}{c}{\textbf{Topicnode}} \\

\cmidrule(lr){2-2}
\cmidrule(lr){3-6}
\cmidrule(lr){8-11}
\cmidrule(lr){12-14}
\cmidrule(lr){15-20}

&
&
\multicolumn{4}{c}{chunk size}

&
&
\multicolumn{4}{c}{chunk size}

& \multicolumn{3}{c}{t}

& \multicolumn{2}{c}{st=0.3}
& \multicolumn{2}{c}{st=0.5}
& \multicolumn{2}{c}{st=0.8} \\

\cmidrule(lr){3-6}
\cmidrule(lr){8-11}
\cmidrule(lr){12-14}
\cmidrule(lr){15-16}
\cmidrule(lr){17-18}
\cmidrule(lr){19-20}

\textbf{LLM}

& 
& 128 & 256 & 512 & 1024

& \textbf{Emb.}

& 128 & 256 & 512 & 1024

& 0.3 & 0.5 & 0.7

& {\scriptsize ws=2} & {\scriptsize ws=5}

& {\scriptsize ws=2} & {\scriptsize ws=5}

& {\scriptsize ws=2} & {\scriptsize ws=5} \\

\midrule

\multirow{6}{*}{\rotatebox[origin=c]{90}{Llama-8B}}*

& \fbox{.713}
& .705 & .712 & .71 & .707
& mxbai
& .701 & .700 & .704 & \textbf{.716}
& .706 & .682 & .676
& \cellcolor{blue!15}\textbf{.734} & \textbf{.724}
& \textbf{.719} & \textbf{.719}
& .646 & .622 \\*

&
&  &  &  & 
& bge
& \cellcolor{blue!15}.703 & \cellcolor{blue!15}\textbf{.714} & \cellcolor{blue!15}\textbf{.727} & \cellcolor{blue!15}\textbf{.720}
& .707 & \cellcolor{blue!15}.692 & \cellcolor{blue!15}.677
& \textbf{.724} & \cellcolor{blue!15}\textbf{.726}
& \textbf{.731} & \textbf{.725}
& \cellcolor{blue!15}.698 & .667 \\*

&
&  &  &  & 
& e5
& .670 & .696 & .713 & \textbf{.716}
& .709 & .670 & .672
& \textbf{\uline{.742}} & .669
& .713 & \cellcolor{blue!15}\textbf{.738}
& .685 & \cellcolor{blue!15}.694 \\*

&
&  &  &  & 
& SFR
& .694 & .696 & .699 & \textbf{.716}
& \textbf{.724} & .679 & .664
& \textbf{.722} & .672
& \cellcolor{blue!15}\textbf{.737} & \textbf{.731}
& .651 & .676 \\*

&
&  &  &  & 
& Qwen3
& .691 & .694 & .707 & .711
& \cellcolor{blue!15}\textbf{.729} & .677 & .662
& .690 & \textbf{.715}
& .668 & .652
& .646 & .650 \\*

&
&  &  &  & 
& Cohere
& .677 & .697 & .705 & .708
& .713 & .671 & .663
& .647 & .696
& .696 & .679
& .668 & .663 \\*

\midrule

\multirow{6}{*}{\rotatebox[origin=c]{90}{Qwen-32B}}

& \fbox{.771}
& .758 & .763 & .763 & .773
& mxbai
& .736 & \cellcolor{blue!15}{.767} & .752 & .757
& \cellcolor{blue!15}{.767} & .730 & .732
& .763 & .747
& .763 & .742
& .718 & .701 \\

&
&  &  &  & 
& bge
& \cellcolor{blue!15}{.750} & .757 & .759 & \cellcolor{blue!15}{.764}
& .761 & \cellcolor{blue!15}{.736} & \cellcolor{blue!15}{.743}
& \cellcolor{blue!15}{.768} & .754
& .756 & .758
& .729 & .714 \\

&
&  &  &  & 
& e5
& .729 & .761 & \cellcolor{blue!15}{.762} & .753
& .754 & .725 & .725
& .758 & \cellcolor{blue!15}{.758}
& .759 & .762
& \cellcolor{blue!15}{.740} & \cellcolor{blue!15}{.740} \\

&
&  &  &  & 
& SFR
& .696 & .722 & .740 & .747
& .762 & \cellcolor{blue!15}{.736} & .720
& .760 & .748
& \cellcolor{blue!15}\textbf{\uline{.774}} & \cellcolor{blue!15}{.764}
& .731 & .726 \\

&
&  &  &  & 
& Qwen3
& .724 & .745 & .738 & .746
& .751 & .724 & .716
& .750 & .748
& .712 & .715
& .705 & .708 \\

&
&  &  &  & 
& Cohere
& .725 & .753 & .757 & .752
& .761 & .712 & .715
& .754 & .746
& .734 & .718
& .710 & .718 \\

\midrule

{\rotatebox[origin=c]{90}{70B}}

& \fbox{\uline{.752}}
& .712 & .731 & .743 & .736
& bge
& .719 & .749 & .743 & .742
& .754 & .715 & .716
& \textbf{\uline{.752}} & .743
& .741 & .738
& .740 & .720 \\

\midrule

{\rotatebox[origin=c]{90}{5.1}}

& \fbox{\textcolor{green!60!black}{\textbf{\uline{.818}}}}
& .796 & .802 & .810 & .806
& bge
& .776 & .814 & .792 & .813
& .803 & .792 & .788
& .814 & .806
& .802 & .805
& .788 & .776 \\

\bottomrule

\end{tabular}

\caption[RD Identification F1 scores across strategies and models]{\textbf{RD Identification} F1 scores across retrieval strategies, hyperparameters, embedding models and LLMs. BM25 does not use embeddings.}

\label{tab:rdi_results_appendix}

\end{table*}

\begin{table*}[t!]

\centering

\scriptsize

\setlength{\tabcolsep}{4pt}

\begin{tabular}{@{}lccccc l cccc ccc cccccc@{}}

\toprule

& \multicolumn{1}{c}{\textbf{LC}}
& \multicolumn{4}{c}{\textbf{BM25}}

&
& \multicolumn{4}{c}{\textbf{Dense}}

& \multicolumn{3}{c}{\textbf{Adaptive}}

& \multicolumn{6}{c}{\textbf{Topicnode}} \\

\cmidrule(lr){2-2}
\cmidrule(lr){3-6}
\cmidrule(lr){8-11}
\cmidrule(lr){12-14}
\cmidrule(lr){15-20}

&
&
\multicolumn{4}{c}{chunk size}

&
&
\multicolumn{4}{c}{chunk size}

& \multicolumn{3}{c}{t}

& \multicolumn{2}{c}{st=0.3}
& \multicolumn{2}{c}{st=0.5}
& \multicolumn{2}{c}{st=0.8} \\

\cmidrule(lr){3-6}
\cmidrule(lr){8-11}
\cmidrule(lr){12-14}
\cmidrule(lr){15-16}
\cmidrule(lr){17-18}
\cmidrule(lr){19-20}

\textbf{LLM}

& 
& 128 & 256 & 512 & 1024

& \textbf{Emb.}

& 128 & 256 & 512 & 1024

& 0.3 & 0.5 & 0.7

& {\scriptsize ws=2} & {\scriptsize ws=5}

& {\scriptsize ws=2} & {\scriptsize ws=5}

& {\scriptsize ws=2} & {\scriptsize ws=5} \\

\midrule

\multirow{6}{*}{\rotatebox[origin=c]{90}{Llama-8B}}*

& \fbox{.414}
& .327 & .382 & \textbf{.443} & .400
& mxbai
& .367 & .353 & .379 & .385
& \textbf{.418} & .362 & \cellcolor{blue!15}{.384}
& .361 & .375
& .359 & .371
& .327 & .319 \\*

&
&  &  &  &
& bge
& .347 & \cellcolor{blue!15}\textbf{.442} & .406 & .404
& \textbf{.424} & \cellcolor{blue!15}.380 & .370
& .353 & .353
& .393 & .394
& \cellcolor{blue!15}.356 & .326 \\*

&
&  &  &  &
& e5
& .284 & .346 & .344 & .371
& .410 & .298 & .298
& .376 & .315
& .342 & .389
& .299 & .312 \\*

&
&  &  &  &
& SFR
& \cellcolor{blue!15}.397 & .399 & \cellcolor{blue!15}.413 & .394
& .413 & .373 & .359
& .381 & \cellcolor{blue!15}.409
& \cellcolor{blue!15}.406 & \cellcolor{blue!15}.406
& .307 & \cellcolor{blue!15}.337 \\*

&
&  &  &  &
& Qwen3
& \cellcolor{blue!15}.397 & .389 & .408 & \cellcolor{blue!15}\textbf{.419}
& \cellcolor{blue!15}\textbf{\uline{.447}} & .369 & .365
& \cellcolor{blue!15}.393 & .379
& .366 & .333
& .328 & .336 \\*

&
&  &  &  &
& Cohere
& .371 & .374 & .410 & .363
& .412 & .339 & .353
& .346 & .376
& .346 & .359
& .319 & .340 \\*

\midrule

\multirow{6}{*}{\rotatebox[origin=c]{90}{Qwen-32B}}

& \fbox{\uline{.605}}
& .487 & .535 & .554 & .557
& mxbai
& .461 & .526 & \cellcolor{blue!15}{.542} & .549
& .576 & .487 & \cellcolor{blue!15}{.496}
& .508 & .504
& \cellcolor{blue!15}{.536} & \cellcolor{blue!15}{.498}
& .431 & .383 \\

&
&  &  &  &
& bge
& .488 & \cellcolor{blue!15}{.542} & .529 & .556
& \cellcolor{blue!15}{.590} & \cellcolor{blue!15}{.496} & \cellcolor{blue!15}{.496}
& .510 & .498
& .497 & .494
& \cellcolor{blue!15}{.450} & .385 \\

&
&  &  &  &
& e5
& .453 & .475 & .529 & .526
& .512 & .439 & .427
& .489 & .480
& .476 & .480
& .411 & .415 \\

&
&  &  &  &
& SFR
& .470 & .502 & .487 & .528
& .501 & .484 & .454
& \cellcolor{blue!15}{.520} & \cellcolor{blue!15}{.510}
& .521 & .479
& .432 & \cellcolor{blue!15}{.469} \\

&
&  &  &  &
& Qwen3
& \cellcolor{blue!15}{.506} & .498 & .534 & \cellcolor{blue!15}{.562}
& .551 & .463 & .467
& .505 & .485
& .453 & .466
& .424 & .434 \\

&
&  &  &  &
& Cohere
& .450 & .524 & .528 & .529
& .531 & .459 & .441
& .490 & .444
& .472 & .447
& .402 & .384 \\

\midrule

{\rotatebox[origin=c]{90}{70B}}

& \fbox{.555}
& .402 & .528 & .536 & \textbf{\uline{.573}}
& bge
& .473 & .542 & .518 & .491
& .553 & .474 & .483
& .517 & .484
& .372 & .406
& .393 & .344 \\

\midrule

{\rotatebox[origin=c]{90}{5.1}}

& \fbox{.608}
& .545 & .594 & \textbf{.618} & .575
& bge
& .517 & .556 & .531 & .527
& .578 & \textbf{.613} & \textcolor{green!60!black}{\textbf{\uline{.62}}}
& .601 & .56
& .481 & .506
& .441 & .444 \\

\bottomrule

\end{tabular}

\caption[RD Assessment F1 scores across strategies and models]{\textbf{RD Assessment} F1 scores across retrieval strategies, hyperparameters, embedding models and LLMs.}

\label{tab:rda_results_appendix}

\end{table*}
\clearpage

\clearpage
\section{ARDTrA Analytical Framework}
\label{sec:appendix_ardtra_fulltext}
\raggedbottom
In this Appendix we present the ARDTrA Analytical Framework, described in Section \ref{sec:dataset:guidelines}. The framework consists of two main tasks: Research Design Identification (RD-I, Section \ref{sec:appendix_rd_identification}) and Research Design Assessment (RD-A, Section \ref{sec:appendix_rd_assessment}).
\subsection{Research Design Identification (6 questions)}
\label{sec:appendix_rd_identification}
\begin{enumerate}
    \item \textit{What type of paper is this?}
    \begin{enumerate}
    \item Empirical - analyzes real-world data to address research questions
    \item Theoretical - develops or discusses conceptual, graphical or mathematical theoretical models
    \item Methodological - develops, discusses or evaluates statistical or research method
    \item Review paper - summarizes or synthesizes other studies (e.g., systematic reviews, meta-analyses)
    \item Opinion paper, comment, reply - offers commentary or responds to another paper
    \item Other
    \end{enumerate}

    \item \textit{Does the paper use quantitative or qualitative research designs and data?}
    \begin{enumerate}
    \item Quantitative research design and data - uses primarily numeric or categorical data (e.g., from surveys, administrative data, experiments) and statistical models
    \item Qualitative research design and data - uses primarily non-numerical data derived from interviews, participant observation, focus groups, narratives or other qualitative and participatory data collections methods
    \item Other
    \end{enumerate}

    \item \textit{Which type of analyses best describes the paper?}
    \begin{enumerate}
        \item Causal Analysis - investigates causal relationships between two or more variables, e.g., investigates whether a variable or intervention causes a change in another
        \item Descriptive Analysis - describes the state, evolution or relationships among variables without making causal claims
        \item Predictive or Simulation Analysis - predicts or simulates the value or evolution of one or multiple variables
        \item Other
    \end{enumerate}

    \item \textit{Does the study use any of the following causal research design/methods: Experimental designs (e.g., field experiments, survey experiments, lab-experiments), Selection-on-observables (e.g., regression adjustment, matching, weighting, doubly robust), Instrumental variables, Threshold-based designs (e.g., regression discontinuity, regression kink, bunching), Difference-in-differences and related designs (e.g., event studies, changes-in-changes, triple differences), Synthetic control method and related designs (e.g., augmented synthetic control, synthetic difference in differences)?}
    \begin{enumerate}
        \item Yes
        \item No
    \end{enumerate}

     \item \textit{Which of the following is the main or preferred research design, associated with the credibility revolution, potential outcome framework or design based approach, used in the paper?}
    \begin{enumerate}
        \item Experimental designs (e.g., field experiments, survey experiments and lab-experiments)
        \item Selection-on-observables (e.g., regression adjustment, matching, weighting, doubly robust)
        \item Instrumental variables
        \item Threshold-based designs (e.g., regression discontinuity, regression kink or bunching)
        \item Difference-in-differences and related designs (e.g., event studies, changes-in-changes, triple differences)
        \item Synthetic control method and related designs (e.g., augmented synthetic control, synthetic difference in differences)
        \item Other
    \end{enumerate}

    \item \textit{Does the paper use additional research designs beyond the selected one, fitting under the design-based causal inference approaches associated with the \enquote{credibility revolution}?}
    \begin{enumerate}
        \item Experimental designs (field experiments, survey experiments and lab-experiments)
        \item Selection-on-observables
        \item Instrumental variables
        \item Threshold-based designs (regression discontinuity, regression kink or bunching)
        \item Difference-in-differences and related designs (i.e., event studies, changes-in-changes, triple differences)
        \item Synthetic control method and related designs (augmented synthetic control, synthetic difference in differences)
    \end{enumerate}
\end{enumerate}

\subsection{Research Design Assessment}
\label{sec:appendix_rd_assessment}

\subsubsection{Experimental Designs (ED)}
\label{sec:experimental_designs}

\begin{enumerate}
    
    \item \textit{What is the randomization strategy used in the study?}
    \begin{enumerate}
        \item Complete (simple) randomization: units are independently assigned to treatment/control with fixed probability
        \item Matched-pair or matched-set randomization: treatment assigned within small, covariate-matched group
        \item Stratified (blocked) randomization: treatment assigned within pre-defined blocks based on key covariates
        \item Covariate-adaptive randomization: assignment probabilities adjusted dynamically to improve balance
        \item Re-randomization: randomization repeated until balance criteria are met
        \item Not reported or described
        \item Other
    \end{enumerate}

    \item \textit{What is the unit of assignment to treatment and control conditions?}
    \begin{enumerate}
        \item Individual level assignment: each unit is independently assigned (e.g., individuals, firms)
        \item Cluster level assignment: units are assigned in groups (e.g., schools, firms, villages)
        \item Multiple levels: assignment occurs at more than one level, common in designs aimed to capture spillover effects (e.g., schools then classes within schools)
        \item Not reported or unclear
        \item Other
    \end{enumerate}

    \item \textit{Was the experimental protocol and analysis:}
    \begin{enumerate}
        \item Pre-registered: a pre-analysis plan or trial registration was filed prior to data collection / analyses
        \item Approved by an ethical board (e.g., IRB)
        \item Not reported or unclear
    \end{enumerate}

    \item \textit{Does the study include any balance / randomization checks?}
    \begin{enumerate}
        \item Pre-determined covariates (e.g., baseline socio-demographic characteristics)
        \item Pre-determined outcomes or related proxies (i.e., baseline outcome variables)
        \item Post-treatment variables not expected to be affected by the treatment (i.e., placebo outcomes)
        \item No balance test reported
        \item Other 
    \end{enumerate}

    \item \textit{Does the study provide evidence of good covariate balance?}
    \begin{enumerate}
        \item Yes - balance is high with only small differences between groups
        \item Partially - balance is mostly good, but some relevant differences remain
        \item No - balance is poor, with substantial differences between groups
        \item Not discussed or not assessed 
    \end{enumerate}

    \item \textit{What is the level of general attrition, in percentage terms? Note: if there are multiple treatment groups with different levels of attrition please select multiple values for each row.}
    \begin{enumerate}
        \item Less than 5 percent
        \item 5-15 percent
        \item 16-25 percent
        \item 26-40 percent
        \item More than 40 percent
        \item Not applicable / relevant (i.e., one-shot data collection and randomization, use of administrative data)
        \item Not reported
    \end{enumerate}

    \item \textit{What is the level of differential attrition, in percentage points? Note: if there are multiple treatment groups with different levels of attrition please select multiple values for each row.}
    \begin{enumerate}
        \item Less than 5 percentage points
        \item 5-15 percentage points
        \item 16-25 percentage points
        \item 26-40 percentage points
        \item More than 40 percentage points
        \item Not applicable / relevant (i.e., one-shot data collection and randomization, use of administrative data)
        \item Not reported
    \end{enumerate}

    \item \textit{What estimation method(s) are used to estimate treatment effects in the main analysis?}
    \begin{enumerate}
        \item Difference-in-means
        \item Regression models (OLS, probit, logit etc)
        \item Matching, weighting or related approaches (e.g., propensity score matching, inverse probability weighting, doubly robust)
        \item ANOVA / ANCOVA
        \item Non-parametric methods (e.g., Wilcoxon Rank-Sum Test, Mann-Whitney U, Kolmogorov-Smirnov Test, Quantile treatment effects, Kruskal-Wallis Test, etc)
        \item Bayesian approaches
        \item Permutation-based test or related (Fisher randomization tests)
        \item Multilevel (Hierarchical) Models (e.g., mixed-effects models, random intercepts/slopes)
        \item Machine learning–based estimators (e.g., lasso, causal forest, double machine learning)
        \item Other
    \end{enumerate}

    \item \textit{Is clustering appropriately accounted for in the inference procedure? Note: The recommended practice is to account for clustering at the level of unit assignment when the unit of analysis is at a lower level than the cluster (e.g., schools assigned but data is at student level). However, if the unit of analysis is at the level of clustering (e.g., schools assigned, and data is aggregated at school level), clustering is not necessary. Clustering is also necessary when there are multiple observations per unit used in the analysis (e.g., repeated measures). Extra: clustering is also necessary when there is a risk of interference or spillovers, but this is rarely done, so if the paper reports such, please choose the  ``g) Other'' option, ignoring all other response options.}
    \begin{enumerate}
        \item Yes - clustering at level of treatment assignment
        \item Yes - clustering below treatment assignment (e.g., at class level when schools are assigned)
        \item Yes - clustering above treatment assignment (e.g., at school level, when classes are assigned)
        \item No - clustering not used and not necessary (e.g., individual-level or clustered assignment but the unit of analysis matches the treatment assignment level)
        \item No - clustering not used, but it should have been\item Unclear / not reported
        \item Other
    \end{enumerate}

    \item \textit{Does the paper report any of the following robustness or sensitivity checks? Select the ``f) Other'' option only if you identify other highly relevant falsification tests for the identification strategy}
    \begin{enumerate}
        \item Sensitivity to the violation of the identification assumption (e.g., bounding, partial identification; i.e., for cases with high attrition rates or implementation issues)
        \item Sensitivity to alternative estimators
        \item Sensitivity to inclusion or exclusion of covariates (e.g., robustness to controls or pre-treatment variables)
        \item Sensitivity to using alternative inference approaches (e.g., different inference method, level of clustering etc.)
        \item Robustness to sample exclusions (e.g., outlier removal, trimming)
        \item Other
    \end{enumerate}

    \item \textit{Does the paper suffer from any of the potential issues?}
    \begin{enumerate}
        \item Analyses include units not randomly assigned (e.g., units entering the program after randomization)
        \item Analyses do not account for differing assignment probabilities (e.g., the probability of treatment varies considerably within strata but the analysis does not account for this)
        \item A randomized unit’s assigned condition is not the same as the unit’s analyzed condition (i.e., mismatch between assigned and analyzed treatment status, e.g., controls non-complying with assignment considered part of the treatment group in the analysis)
        \item Exclusion of randomized units based on post-assignment criteria (e.g., excluding treatment units not taking up the treatment)
        \item No, none of the above
    \end{enumerate}
\end{enumerate}

\subsubsection{Threshold-based designs (e.g., regression discontinuity, regression kink or bunching)}
\label{sec:threshold_based_designs}
\begin{enumerate}
    
    \item \textit{What RDD framework does the paper use in its main analysis?}
    \begin{enumerate}
        \item Continuity-based framework - assumes potential outcomes are continuous at the cutoff (the most frequent approach)
        \item Local randomization framework - treats units near the cutoff as randomly assigned to treatment/control (often with discrete running variables or small sample sizes, or as robustness check to the continuity approach. It relies on stronger assumptions.)
        \item Other
    \end{enumerate}

    \item \textit{Which of the following RDD features best describes the paper?}
    \begin{enumerate}
        \item Single running variable
        \item Single cutoff
        \item Multiple cutoffs
        \item Multiple running variables (i.e., geographical RDDs, scholarship assigned based on cutoffs in both language and math tests)
        \item Other
    \end{enumerate}

    \item \textit{Is the running variable approximately continuous or discrete?}
    \begin{enumerate}
        \item Approximately continuous: there is a large number of values on each side of the cutoff
        \item Discrete with more than 10 unique values on each side of the cutoff
        \item Discrete with 10 or fewer unique values on each side of the cutoff
        \item Other
    \end{enumerate}

     \item \textit{Does the paper provide evidence for the continuity of the running variable density around the cutoff?}
    \begin{enumerate}
        \item Histograms or similar graphical analysis
        \item Bernoulli or binomial tests around the cutoff (e.g., Cattaneo et al., 2017)
        \item McCrary (2008) local polynomial density estimation
        \item Local polynomial density estimation (e.g., Cattaneo, Jansson, and Ma, 2020)
        \item No specific details provided in the text other than support for the hypothesis (performed in the appendix)
        \item Other
        \item Not provided
    \end{enumerate}

     \item \textit{Does the density continuity test support the identification assumption (no manipulation of the running variable)?}
    \begin{enumerate}
        \item Yes - no strong evidence of manipulation at the cutoff
        \item No - evidence suggests manipulation or discontinuity
        \item Not provided
    \end{enumerate}

     \item \textit{What estimation approach is used to obtain point estimates?}
    \begin{enumerate}
        \item Global polynomial methods, i.e., regression using observations over the entire support of the running variable, often including higher order polynomials
        \item Local polynomial methods, i.e., regression using only observations close to the cutoff, generally with low order polynomials
        \item Randomization inference within a window close to the cutoff (Fisherian approach)
        \item Ordinary least squares or similar large sample approaches in the local randomization framework (based on Neyman or superpopulation approaches)
        \item Other
    \end{enumerate}

     \item \textit{How was the bandwidth (in continuity-based RDD) or window (in local randomization RDD) selected?}
    \begin{enumerate}
        \item MSE-optimal bandwidth - e.g., from Calonico et al. (2014), based on mean squared error minimization
        \item Covariate-based window selection (local randomization framework)
        \item Manual selection
        \item Cross-validation (e.g., as in Ludwig and Miller, 2007)
        \item IK algorithm for optimal bandwidth selector proposed by Imbens and Kalyanaraman (2012)
        \item Other
    \end{enumerate}

     \item \textit{Is clustering appropriately accounted for in the inference procedure? Note: The recommended practice is to account for clustering at the level of unit assignment when the unit of analysis is at a lower level than the cluster (e.g., schools assigned but data is at student level). However, if the unit of analysis is at the level of clustering (e.g., schools assigned, and data is aggregated at school level), clustering is not necessary. Clustering is also necessary when there are multiple observations per unit used in the analysis (e.g., repeated measures, panel data). Extra: clustering is also necessary when there is a risk of interference or spillovers, but this is rarely done, so if the paper reports such, please indicate in other, ignoring all other response options.}
    \begin{enumerate}
        \item Yes - clustering at level of treatment assignment
        \item Yes - clustering below treatment assignment (e.g., at class level when schools are assigned)
        \item Yes - clustering above treatment assignment (e.g., at school level, when classes are assigned)
        \item No - clustering not used and not necessary (e.g., individual-level or clustered assignment but the unit of analysis matches the treatment assignment level)
        \item No - clustering not used, but it should have been
        \item Unclear / not reported
        \item Other
    \end{enumerate}

    \item \textit{Does the paper provide any of the following falsification (e.g., balance, placebo) tests? Please select the \"e) Other\" option only if it is a highly relevant falsification test for the identification strategy}
    \begin{enumerate}
        \item Pre-determined covariate or outcome balance test - checking continuity or balance of predetermined covariates at the cutoff
        \item Placebo outcomes test - applying the same design / analysis replacing the outcome variable with one which should not be impacted by the treatment
        \item Placebo population test - applying the same design / analysis in a sample where the treatment is expected to not have an effect
        \item Placebo treatment - applying the same design / analysis replacing the treatment variable with one that should not affect the outcome. In the case of the RD this is usually done by using multiple placebo cutoffs. If something different was done, please indicate in the comments.
        \item Other
    \end{enumerate}

    \item \textit{Does the paper provide any of the following sensitivity or robustness checks? Please select the ``h) Other'' option only in the case of highly relevant sensitivity tests for the identification strategy}
    \begin{enumerate}
        \item Sensitivity to the violation of the identification assumptions (e.g., bounding, partial identification)
        \item Sensitivity to the exclusion of observations near the cutoff (the so-called donut hole approach)
        \item Sensitivity to bandwidth or window choices (usually making it smaller, thus excluding observations away from the cutoff)
        \item Sensitivity to using an alternative RDD framework (i.e., local randomization vs continuity)
        \item Robustness to alternative kernels or polynomial orders
        \item Sensitivity to the inclusion or exclusion of covariates (robustness to controlling for pre-treatment variables)
        \item Sensitivity to using alternative inference approaches (e.g., different inference method, level of clustering etc.)
        \item Other
    \end{enumerate}
\end{enumerate}

\subsubsection{Instrumental Variables}
\label{sec:instrumental_variables}
\begin{enumerate}
    \item \textit{How is the plausibility of the identification assumption supported in the paper? In an instrumental variables design, the key identification assumption is that the instrument is independent of unobserved determinants of the outcome — either unconditionally or conditional on observed covariates. This is often not guaranteed by design, especially in observational studies, and must be defended through theoretical arguments, institutional details, balance tests, placebo tests, or robustness checks. Being able to argue convincingly that the instrument is as good as randomly assigned can help strengthen the credibility of the IV strategy. You may select more than one option if the paper combines multiple forms of justification (e.g., uses empirical tests and argues for quasi-random variation). Consider both what the authors claim and the kind of variation they exploit. Do not infer justification that is not discussed or implied by the paper.}
    \begin{enumerate}
        \item Through arguments supporting a quasi-random assignment due to plausibly exogenous policy variation. For example: assignment is determined by administrative quirks, implementation delays, pilot rollout, or other arbitrary features of policy design that are plausibly unrelated to potential outcome levels and trends. This often corresponds to the so-called natural experiments due to policy variations.
        \item Through other plausibly exogenous variation driven by contextual or historical factors. Assignment is determined by contextual forces such as natural disasters, conflict, migration patterns, geographical features or long-standing institutional or historical differences, claimed to generate quasi-random variation. This often corresponds to the so-called natural experiments due to contextual factors.
        \item Through policy eligibility rules that allow comparison between eligible and non-eligible units. This applies when the policy is assigned based on fixed rules - such as socio-demographic characteristics (e.g., age, income, family status) or geography (e.g., some regions treated, others not) - and units cannot self-select into treatment (or would have to incur substantial costs to do so). This justification is common in difference-in-differences designs.
        \item Credibility defended based on theoretical or substantive reasoning (e.g., claims based on social science theories and limited institutional, external or historical variation)
        \item Through reference to previous studies using similar designs without further justification that the design is valid in the current setting.
        \item Identification defended through empirical econometric tests (e.g., the paper supports its identification strategy through balance checks, placebo tests, sensitivity checks etc.)
        \item Other
    \end{enumerate}

    \item \textit{What type of instrument are used in the study?}
    \begin{enumerate}
        \item Randomized assignment - instrument is based on a randomized experiment (e.g., encouragement design, randomized eligibility)
        \item Rules or policy changes - instrument arises from policy thresholds or quasi-random eligibility criteria (e.g., as in RDD settings)
        \item Examiner designs (e.g., judge leniency, using disparities among judges and other decision makers to identify causal effects)
        \item Geography, weather, or climate-based - instrument relies on spatial or natural variation (e.g., rainfall, elevation, proximity to coast or borders)
        \item Historical variation - instrument is based on long-past events or historical institutions (e.g., colonial presence, migration history)
        \item Treatment diffusion (e.g., using US aid in neighboring countries as instrument for US aid in the country analyzed)
        \item Shift-share / Bartik instruments
        \item Other
    \end{enumerate}

    \item \textit{How many instruments are used in the main analysis?}
    \begin{enumerate}
         \item One instrument
         \item Two Instruments
         \item 3-10 instruments
         \item 11-20 instruments
         \item More than 20 instruments
         \item Unclear
    \end{enumerate}

    \item \textit{What is the first-stage F-statistic in the preferred specifications? If there is no clearly stated preferred specification, select all relevant values from the main tables or text.}
    \begin{enumerate}
        \item Not reported, but authors claim the instrument is strong
        \item Not reported, but authors acknowledge the instrument is weak
        \item Less than 10
        \item 10-20
        \item 21-30
        \item 31-50
        \item 51-100
        \item 101-200
        \item Over 200
        \item Not reported and not possible to assess instrument validity
        \item Not reported but can be calculated based on information report to the reader, such as the first stage t-test
        \item Other
    \end{enumerate}

    \item \textit{What type of F test statistic is computed?}
    \begin{enumerate}
        \item Not discussed or unclear
        \item Standard First-Stage F-Statistic assuming homoskedasticity
        \item Kleibergen-Paap Wald F-Statistic (Heteroskedastic and Clustered Robust)
        \item Effective F-Statistic (Montiel-Olea and Pflueger)
        \item Cragg-Donald F-Statistic
        \item Sanderson-Windmeijer (SW) Conditional F-Statistic
        \item Bootstrapped F-statistic
        \item Other
    \end{enumerate}

    \item \textit{Which estimator(s) does the study use to estimate the causal effect in the main analysis?}
    \begin{enumerate}
        \item Two-stage Least Squares (2SLS), including Wald estimator (Wald 1940, Angrist and Krueger 1991)
        \item Limited Information Maximum Likelihood (LIML) (Anderson and Rubin, 1949)
        \item Bias-Corrected 2SLS (Nagar, 1959,  Donald and Newey, 2001, Kolesár et al., 2015, Anatolyev, 2011)
        \item Jackknife IV (JIVE), sometimes referred to as “leave-one-out estimator” (Angrist, Imbens and Krueger, 1999)
        \item Continuously Updated GMM (CUE) (Hansen, Heaton, and Yaron, 1996)
        \item Standard two-step GMM (Generalized Method of Moments)
        \item Local Instrumental Variables (LIV) (Heckman and Vytlacil, 1999)
        \item Unclear or difficult to assess
        \item Other
    \end{enumerate}

    \item \textit{Inference is performed through:}
    \begin{enumerate}
        \item Standard IV asymptotic inference
        \item Bootstrap-based Inference
        \item Anderson-Rubin (AR) Test
        \item Conditional Likelihood Ratio (CLR) Test (e.g., Moreira, 2003)
        \item tF procedure (Lee et al. 2022)
        \item Unclear or difficult to assess
        \item Other
    \end{enumerate}

    \item \textit{Is clustering appropriately accounted for in the inference procedure in the second stage? Note: The recommended practice is to account for clustering at the level of unit assignment when the unit of analysis is at a lower level than the cluster (e.g., schools assigned but data is at student level). However, if the unit of analysis is at the level of clustering (e.g., schools assigned, and data is aggregated at school level), clustering is not necessary. Clustering is also necessary when there are multiple observations per unit used in the analysis (e.g., repeated measures, panel data). Extra: clustering is also necessary when there is a risk of interference or spillovers, but this is rarely done, so if the paper reports such, please select  \enquote{g) Other}, ignoring all other response options}"
    \begin{enumerate}
        \item Yes - clustering at level of treatment assignment
        \item Yes - clustering below treatment assignment (e.g., at class level when schools are assigned)
        \item Yes - clustering above treatment assignment (e.g., at school level, when classes are assigned)
        \item No - clustering not used and not necessary (e.g., individual-level or clustered assignment but the unit of analysis matches the treatment assignment level)
        \item No - clustering not used, but it should have been
        \item Unclear / not reported
        \item Other
    \end{enumerate}

    \item \textit{Are the first stage coefficients in line with theoretical predictions?}
    \begin{enumerate}
        \item Yes
        \item No
        \item Not discussed or first stage estimates not reported
        \item Unclear or difficult to assess
    \end{enumerate}

    \item \textit{Does the paper compare the IV estimates to the OLS estimates?}
    \begin{enumerate}
        \item Yes - and IV estimates are larger in absolute values
        \item Yes - and IV estimates are smaller in absolute values
        \item Yes - IV estimates are similar to OLS estimates
        \item No - OLS estimates are not reported
        \item Unclear or difficult to assess
    \end{enumerate}

    \item \textit{If yes, how are the differences between IV and OLS estimates justified?}
    \begin{enumerate}
        \item Selection arguments (or lack of selection bias) - OLS is considered biased due to negative or positive selection into treatment of units
        \item Complier difference - it argues that differences are likely to differences of compliers relative to the population
        \item Measurement error (i.e., downward bias OLS estimates due to measurement error in the treatment variable)
        \item No clear justification offered 
    \end{enumerate}

    \item \textit{Does the study include any falsification (placebo) tests to support the identification strategy? Please select the \enquote{e) Other} option only if it is a highly relevant falsification tests for the identification strategy}
    \begin{enumerate}
        \item Post-treatment placebo outcomes test - applying the same design / analysis replacing the outcome variable with one which should not be impacted by the treatment or instrument
        \item Placebo treatment / instrument - applying the same design / analysis replacing the treatment or instrument variable with one that should not affect the outcome.
        \item Placebo population test - applying the same design / analysis in a sample where the treatment or instrument is expected to not have an effect
        \item Placebo / Balance on pre-treatment covariates or pre-treatment outcomes, either unconditionally or conditional on a sub-set of covariates which are considered sufficient to make the instrument as good as random
        \item Other
    \end{enumerate}

    \item \textit{What type of sensitivity analyses does the study conduct? Please select the \enquote{f) Other} option only in the case of other highly relevant sensitivity tests for the identification strategy}
    \begin{enumerate}
        \item Sensitivity to the violation of the identification assumptions (e.g., bounding, partial identification)
        \item Robustness to alternative estimators
        \item Sensitivity to using alternative inference approaches, first or second stage
        \item Robustness to weak instruments procedures
        \item Robustness to the exclusion of outliers
        \item Other
    \end{enumerate}

    \item \textit{Does the study perform any overidentification tests? Examples of overidentification tests: Sargan test, Hansen J-test}
    \begin{enumerate}
        \item Yes
        \item No overidentification test reported
        \item Not applicable - model is exactly identified (same number of instruments as endogenous variables)
    \end{enumerate}

    \item \textit{Does the paper explicitly discuss or provide formal empirical support for the exclusion restriction in the IV setting? The exclusion restriction assumes that the instrument affects the outcome only through its effect on the treatment (i.e., there is no direct effect of the instrument on the outcome)}
    \begin{enumerate}
        \item Discussion, i.e., discussed qualitatively or formal tests (e.g., supportive empirical evidence) provided
        \item No - does not mention or address
        \item Unclear or hard to assess
    \end{enumerate}

    \item \textit{Does the paper explicitly discuss or provide formal empirical support for the monotonicity assumption in the IV setting? Monotonicity assumes that the instrument affects the treatment in the same direction for all units (i.e., there are no defiers)}
    \begin{enumerate}
        \item Discussion, i.e., discussed qualitatively or formal tests (e.g., supportive empirical evidence) provided
        \item No - does not mention or address
        \item Unclear or hard to assess
    \end{enumerate}
\end{enumerate}

\subsubsection{Selection-on-observables (e.g., regression adjustment, matching, weighting doubly robust}

\begin{enumerate}
    \item \textit{How is the plausibility of the identification assumption supported in the paper?}
    \begin{enumerate}
        \item Through arguments supporting a quasi-random assignment due to plausibly exogenous policy variation. For example: assignment is determined by administrative quirks, implementation delays, pilot rollout, or other arbitrary features of policy design that are plausibly unrelated to potential outcome levels and trends. This often corresponds to the so-called \enquote{natural experiments} due to policy variations.
        \item Through other plausibly exogenous variation driven by contextual or historical factors. Assignment is determined by contextual forces such as natural disasters, conflict, migration patterns, geographical features or long-standing institutional or historical differences, claimed to generate quasi-random variation. This often corresponds to the so-called \enquote{natural experiments} due to contextual factors
        \item Through policy eligibility rules that allow comparison between eligible and non-eligible units This applies when the policy is assigned based on fixed rules — such as socio-demographic characteristics (e.g., age, income, family status) or geography (e.g., some regions treated, others not) - and units cannot self-select into treatment (or would have to incur substantial costs to do so). This justification is common in difference-in-differences designs
        \item Credibility defended based on theoretical or substantive reasoning (e.g., claims based on social science theories and limited institutional, external or historical variation)
        \item Through reference to previous studies using similar designs without further justification that the design is valid in the current setting.
        \item Identification defended through empirical econometric tests (e.g., the paper supports its identification strategy through balance checks, placebo tests, sensitivity checks etc.)
        \item Other
    \end{enumerate}

    \item \textit{Which of the following describe the set of covariates used to support the unconfoundedness assumption?}
    \begin{enumerate}
        \item Basic set of socio-demographic covariates (e.g., gender, age in a study on individuals)
        \item Rich set of socio-demographic covariates (e.g., family education, income, employment status etc.)
        \item Pre-treatment outcome or close proxy available for one period
        \item Pre-treatment outcome or close proxy available for multiple periods
        \item Post-treatment variables potentially impacted by the treatment (i.e., problematic, collider bias risk)
        \item Other
    \end{enumerate}

    \item \textit{Does the study provide any evidence for the credibility of the common support assumption, either on covariates or a distance measure (e.g., propensity score)?}
    \begin{enumerate}
        \item Yes
        \item No
    \end{enumerate}

    \item \textit{Is any trimming or sample restriction applied to improve overlap/common support?}
    \begin{enumerate}
        \item Yes, keeping units on the common support
        \item Yes, dropping units with high or low propensity scores using \textit{ad hoc} thresholds (e.g., 0.1 and 0.9)
        \item Yes, dropping units with high or low propensity scores in a data driven way
        \item Yes, dropping units based on other \textit{ad hoc} criteria
        \item There were no common support issues
        \item No trimming discussed
        \item Other 
    \end{enumerate}

    \item \textit{Does the study provide evidence of good covariate balance after adjustment? Balance may be assessed based on standardized mean differences and variance ratios, plots, or other tables with estimates of differences by group}
    \begin{enumerate}
        \item Yes, balance is high with small remaining differences
        \item Partially, balance is good with some remaining differences in relevant variables
        \item Inadequate balance, with many remaining differences between the groups
        \item Unclear or difficult to assess 
    \end{enumerate}

    \item \textit{Which estimation methods does the study use in the main analyses? Please note that some approaches are combined with regression adjustments. In that case, please select more than one option}
    \begin{enumerate}
        \item Outcome modelling, i.e., regression adjustments (e.g., linear regression, kernel regression, spline-based adjustments)
        \item Exact matching or Coarsened Exact Matching
        \item Matching on the propensity score (nearest neighbor, genetic matching, subclassification etc.)
        \item Weighting methods (e.g., inverse probability weighting [IPW], entropy balancing)
        \item Doubly robust methods (e.g., augmented inverse probability weighting, double machine learning)
        \item Other
    \end{enumerate}

    \item \textit{Is clustering appropriately accounted for in the inference procedure? Note: The recommended practice is to account for clustering at the level of unit assignment when the unit of analysis is at a lower level than the cluster (e.g., schools assigned but data is at student level). However, if the unit of analysis is at the level of clustering (e.g., schools assigned, and data is aggregated at school level), clustering is not necessary. Clustering is also necessary when there are multiple observations per unit used in the analysis (e.g., repeated measures, panel data). Extra: clustering is also necessary when there is a risk of interference or spillovers, but this is rarely done, so if the paper reports such, please indicate in other, ignoring all other response options}
    \begin{enumerate}
        \item Yes - clustering at level of treatment assignment
        \item Yes - clustering below treatment assignment (e.g., at class level when schools are assigned)
        \item Yes - clustering above treatment assignment (e.g., at school level, when classes are assigned)
        \item No - clustering not used and not necessary (e.g., individual-level or clustered assignment but the unit of analysis matches the treatment assignment level)
        \item No - clustering not used, but it should have been
        \item Unclear / not reported
        \item Other 
    \end{enumerate}

    \item \textit{Does the study include any falsification (placebo) tests to support the identification strategy? Please select the ``f) Other'' option only if it is a highly relevant falsification test for the identification strategy}
    \begin{enumerate}
        \item Post-treatment placebo outcomes test - applying the same design / analysis replacing the outcome variable with one which should not be impacted by the treatment
        \item Placebo treatment - applying the same design / analysis replacing the treatment variable with one that should not affect the outcome.
        \item Placebo population test - applying the same design / analysis in a sample where the treatment is expected to not have an effect
        \item Placebo / Balance on pre-treatment covariates - excluding a sub-set of covariates from the balancing and checking for differences
        \item Placebo / Balance on Placebo pre-treatment outcomes or close proxies - e.g., excluding one or more pre-treatment outcomes from the balancing and assessing for differences
        \item Other
    \end{enumerate}

    \item \textit{What type of sensitivity analyses does the study conduct?}
    \begin{enumerate}
        \item Sensitivity to the violation of the identification assumptions (e.g., bounding, partial identification)
        \item Robustness to alternative estimators
        \item Sensitivity to the selection of covariates included in the balancing
        \item Robustness to alternative trimming strategies
        \item Sensitivity to using alternative inference approaches (e.g., different inference method, level of clustering etc.)
        \item Other
    \end{enumerate}
\end{enumerate}

\subsubsection{Difference in Differences (DiD) and related designs (e.g., event studies, changes-in-changes, triple differences)}

\begin{enumerate}
    \item \textit{How is the credibility of main identification strategy supported in the paper? In a DiD setting the key identification assumption is that of parallel trends. This is generally defended empirically by showing that there were parallel trends before the treatment and assuming that such trends are assumed to hold also in the post-treatment period in the absence of the treatment. This is not always credible, as assignment may be due to expectations regarding the evolution of the outcome in the post-treatment period (e.g., there are expectations that growth will slow down only in some regions and the policy is implemented to contrast that). Being able to argue that this was not the case and that there was some idiosyncratic variation can help strengthen the identification strategy. You may select more than one option if the paper combines multiple forms of justification (e.g., uses empirical tests and argues for quasi-random variation). Consider both what the authors claim and the kind of variation they exploit. Do not infer justification that is not discussed or implied by the paper.}
    \begin{enumerate}
        \item Through arguments supporting a quasi-random assignment due to plausibly exogenous policy variation. For example: assignment is determined by administrative quirks, implementation delays, pilot rollout, or other arbitrary features of policy design that are plausibly unrelated to potential outcome levels and trends. This often corresponds to the so-called ``natural experiments'' due to policy variations.
        \item Through other plausibly exogenous variation driven by contextual or historical factors. Assignment is determined by contextual forces such as natural disasters, conflict, migration patterns, geographical features or long-standing institutional or historical differences, claimed to generate quasi-random variation. This often corresponds to the so-called ``natural experiments'' due to contextual factors
        \item Through policy eligibility rules that allow comparison between eligible and non-eligible units. This applies when the policy is assigned based on fixed rules — such as socio-demographic characteristics (e.g., age, income, family status) or geography (e.g., some regions treated, others not) — and units cannot self-select into treatment (or would have to incur substantial costs to do so). This justification is common in difference-in-differences designs
        \item Credibility defended based on theoretical or substantive reasoning (e.g., claims based on social science theories and limited institutional, external or historical variation)
        \item Through reference to previous studies using similar designs without further justification that the design is valid in the current setting.
        \item Identification defended through empirical econometric tests (e.g., The paper supports its identification strategy through balance checks, placebo tests, sensitivity checks etc.)
        \item Other
    \end{enumerate}

    \item \textit{What is the structure of treatment assignment across units and over time?}
    \begin{enumerate}
        \item Blocked design - all treated units receive treatment at the same time (e.g., single policy shock or eligibility event)
        \item Staggered design - treatment is rolled out at different times across units/groups (e.g., cohort or regional rollouts of the program across time)
        \item Other
    \end{enumerate}

    \item \textit{Is the no-anticipation assumption discussed or addressed? This refers to whether the paper considers whether units could respond before the official treatment start}
    \begin{enumerate}
        \item Yes - discussed and addressed where relevant (e.g., by backdating treatment or showing no early effects)
        \item Not discussed, and anticipation is plausible (e.g., policy announcement precedes implementation)
        \item Not discussed but plausibly there are no anticipation concerns
    \end{enumerate}

    \item \textit{What is the time unit of observation (i.e., data frequency)?}
    \begin{enumerate}
        \item Daily
        \item Weekly
        \item Monthly
        \item Quarterly
        \item Yearly
        \item Other
    \end{enumerate}

    \item \textit{What is the number of pre-periods available (i.e., before the treatment/intervention starting date)? Please note that if the paper argues there are anticipation effects, the treatment starting date should be back-dated thus the number of pre-periods should only consider the period preceding the back-dated treatment starting date.}
    \begin{enumerate}
        \item 1 period
        \item 2-3 periods
        \item 4-5 periods
        \item 6-10 periods
        \item 11-20 periods
        \item 21-40 periods
        \item more than 40 periods
        \item Unclear or difficult to assess based on the information provided
    \end{enumerate}

    \item \textit{What is the number of pre-periods available (i.e., before the treatment/intervention starting date)? Please note that if the paper argues there are anticipation effects, the treatment starting date should be back-dated thus the number of pre-periods should only consider the period preceding the back-dated treatment starting date.}
    \begin{enumerate}
        \item 1 period
        \item 2-3 periods
        \item 4-5 periods
        \item 6-10 periods
        \item 11-20 periods
        \item 21-40 periods
        \item more than 40 periods
        \item Unclear or difficult to assess based on the information provided
    \end{enumerate}

    \item \textit{What is the number of post-periods available (i.e., after the treatment/intervention starting date)? Please note that if the paper argues there are anticipation effects, the treatment starting date should be back-dated thus the number of post-periods should consider the periods after the back-dated treatment starting date}
    \begin{enumerate}
        \item 1 period
        \item 2-3 periods
        \item 4-5 periods
        \item 6-10 periods
        \item 11-20 periods
        \item 21-40 periods
        \item more than 40 periods
        \item Unclear or difficult to assess based on the information provided
    \end{enumerate}

    \item \textit{Does the study provide formal or graphical analysis of pre-treatment trends by treatment groups? Please note that if the paper argues there are anticipation effects, the treatment starting date should be back-dated and such parallel pre-trends analyses should be done on the period preceeding any anticipation effects.}
    \begin{enumerate}
        \item Yes, group-level mean outcome plots over time (e.g., raw or smoothed trends by treatment and control groups)
        \item Yes, event study plots (or estimates reported in a different format such as a table)
        \item Not provided in the main text but argued that parallel trends hold in the pre-treatment period
        \item No - neither discussed nor shown
        \item Other
    \end{enumerate}

    \item \textit{Do the treatment and control groups appear to follow parallel trends prior to the intervention?}
    \begin{enumerate}
        \item Yes - trends are visually and/or statistically similar
        \item No - trends are systematically different / diverging or there are many periods where differences are large
        \item Unclear or difficult to assess based on the information provided
    \end{enumerate}

    \item \textit{Which estimator(s) does the study use in the main analysis to estimate treatment effects?}
    \begin{enumerate}
        \item Standard DiD / two-way fixed effects - two-way fixed effects with group and time fixed effects and a treatment indicator (e.g., classic 2×2 DiD or panel DiD)
        \item Event-study with TWFE - the treatment is typically interacted with relative time periods (e.g., event time indicators) to estimate dynamic effects
        \item Callaway and Sant’Anna (2021) estimator - group-time average treatment effects estimator
        \item Imputation-based estimators - e.g., Borusyak, Jaravel, Spiess (2021), Gardner (2021), Liu et al. (2022), Wooldridge (2022)
        \item Sun and Abraham (2021) - interaction-weighted estimator for dynamic effects under staggered adoption
        \item Stacked DID (i.e., Wing, Freedman and Hollingsworth, 2024)
        \item Other
    \end{enumerate}

    \item \textit{Is clustering appropriately accounted for in the inference procedure? Note: The recommended practice is to account for clustering at the level of unit assignment when the unit of analysis is at a lower level than the cluster (e.g., schools assigned but data is at student level). However, if the unit of analysis is at the level of clustering (e.g., schools assigned, and data is aggregated at school level), clustering is not necessary. Clustering is also necessary when there are multiple observations per unit used in the analysis (e.g., repeated measures, panel data). Extra: clustering is also necessary when there is a risk of interference or spillovers, but this is rarely done, so if the paper reports such, please select \enquote{g) Other}, ignoring all other response options}
    \begin{enumerate}
        \item Yes - clustering at level of treatment assignment
        \item Yes - clustering below treatment assignment (e.g., at class level when schools are assigned)
        \item Yes - clustering above treatment assignment (e.g., at school level, when classes are assigned)
        \item No - clustering not used and not necessary (e.g., individual-level or clustered assignment but the unit of analysis matches the treatment assignment level)
        \item No - clustering not used, but it should have been
        \item Unclear / not reported
        \item Other
    \end{enumerate}

    \item \textit{Does the study include other falsification (placebo) tests? Please select the \enquote{d) Other} option only in the case of highly relevant falsification tests for the identification strategy}
    \begin{enumerate}
        \item Post-treatment placebo outcomes test - applying the same design / analysis replacing the outcome variable with one which should not be impacted by the treatment
        \item Placebo population test - applying the same design / analysis in a sample where the treatment is expected to not have an effect
        \item Placebo treatment - applying the same design / analysis replacing the treatment variable with one that should not affect the outcome.
        \item Other
    \end{enumerate}

    \item \textit{What type of sensitivity analyses does the study conduct? Please select the \enquote{g) Other} option only in the case of highly relevant sensitivity tests for the identification strategy}
    \begin{enumerate}
        \item Sensitivity to the violation of the identification assumptions (e.g., bounding, partial identification)
        \item Sensitivity of the parallel trends assumption to the chosen functional form of the outcome
        \item Assessment of the power of pre-trends tests against economically relevant violations of parallel trends
        \item Robustness to alternative estimators (e.g., using modern estimators are used only as robustness please specify here)
        \item Robustness to assuming parallel trends only conditional on covariates
        \item Sensitivity to using alternative inference approaches (e.g., different inference method, level of clustering etc.)
        \item Other
    \end{enumerate}

    \item \textit{Do any of the following issues or assumptions arise in the study’s treatment effect estimation?}
    \begin{enumerate}
        \item The study includes in the analysis also periods where all units are treated
        \item The study provides theoretical, graphical or formal arguments to argue for the homogeneity of treatment effects (e.g., effects not evolving dynamically, and the same effects across units)
        \item None of the above
    \end{enumerate}   
\end{enumerate}

\subsubsection{Synthetic Control Methods (SCMs) and related designs (e.g., augmented synthetic control, synthetic difference-in-differences)}

\begin{enumerate}
    \item \textit{How is the plausibility of the identification assumption supported in the paper? Select the option that best describes how the study justifies the exogeneity of treatment or instrument assignment. Consider both the source of variation and the justification provided. In a SC setting the implicit assumption is similar to a DiD setting, that is, in the absence of the treatment the trends observed pre-intervention are assumed to hold. This is not always credible, as assignment may be due to expectations regarding the evolution of the outcome in the post-treatment period (e.g., there are expectations that growth will slow down only in some regions and the policy is implemented to contrast that). Being able to argue that this was not the case and that there was some idiosyncratic variation can help strengthen the identification strategy}
    \begin{enumerate}
        \item Through arguments supporting a quasi-random assignment due to plausibly exogenous policy variation. For example: assignment is determined by administrative quirks, implementation delays, pilot rollout, or other arbitrary features of policy design that are plausibly unrelated to potential outcome levels and trends. This often corresponds to the so-called ``natural experiments'' due to policy variations.
        \item Through other plausibly exogenous variation driven by contextual or historical factors. Assignment is determined by contextual forces such as natural disasters, conflict, migration patterns, geographical features or long-standing institutional or historical differences, claimed to generate quasi-random variation. This often corresponds to the so-called ``natural experiments'' due to contextual factors
        \item Through policy eligibility rules that allow comparison between eligible and non-eligible units. This applies when the policy is assigned based on fixed rules — such as socio-demographic characteristics (e.g., age, income, family status) or geography (e.g., some regions treated, others not) — and units cannot self-select into treatment (or would have to incur substantial costs to do so). This justification is common in difference-in-differences designs
        \item Credibility defended based on theoretical or substantive reasoning (e.g., claims based on social science theories and limited institutional, external or historical variation)
        \item Through reference to previous studies using similar designs without further justification that the design is valid in the current setting.
        \item Identification defended through empirical econometric tests (e.g., the paper supports its identification strategy through balance checks, placebo tests, sensitivity checks etc.)
        \item Other
    \end{enumerate}

    \item \textit{Please indicate the number of treated units used in the study}
    \begin{enumerate}
        \item 1
        \item 2-5
        \item 6-20
        \item 21-50
        \item 51-100
        \item More than 100
    \end{enumerate}

    \item \textit{Please indicate the number of control units used in the study}
    \begin{enumerate}
        \item 1
        \item 2-5
        \item 6-20
        \item 21-50
        \item 51-100
        \item More than 100
    \end{enumerate}

    \item \textit{Is the no-anticipation assumption discussed or addressed? This refers to whether the paper considers whether units could respond before the official treatment start.}
    \begin{enumerate}
        \item Yes - discussed and addressed where relevant (e.g., by backdating treatment or showing no early effects)
        \item Not discussed, and anticipation is plausible (e.g., policy announcement precedes implementation)
        \item Not discussed but plausibly there are no anticipation concerns
    \end{enumerate}

    \item \textit{What is the time unit of observation (i.e., data frequency)?}
    \begin{enumerate}
        \item Daily
        \item Weekly
        \item Monthly
        \item Quarterly
        \item Yearly
        \item Other
    \end{enumerate}

    \item \textit{What is the number of pre-periods available (i.e., before the treatment/intervention starting date)? Please note that if the paper argues there are anticipation effects, the treatment starting date should be back-dated thus the number of pre-periods should only consider the period preceding the back-dated treatment starting date.}
    \begin{enumerate}
        \item 1 period
        \item 2-3 periods
        \item 4-5 periods
        \item 6-10 periods
        \item 11-20 periods
        \item 21-40 periods
        \item More than 40 periods
        \item Unclear or difficult to assess based on the information provided
    \end{enumerate}

    \item \textit{What is the number of post-periods available (i.e., after the treatment/intervention starting date)? Please note that if the paper argues there are anticipation effects, the treatment starting date should be back-dated thus the number of post-periods should consider the periods after the back-dated treatment starting date}
    \begin{enumerate}
        \item 1 period
        \item 2-3 periods
        \item 4-5 periods
        \item 6-10 periods
        \item 11-20 periods
        \item 21-40 periods
        \item More than 40 periods
        \item Unclear or difficult to assess based on the information provided
    \end{enumerate}

    \item \textit{Is there a high volatility of the outcome in the pre-treatment period (or average outcome in the case of multiple units)? Outcome variables that include substantial random noise elevate the risk of over-fitting. It is advisable to remove it via filtering, in both the exposed unit as well as in the units in the donor pool, before applying synthetic control techniques. This is especially concerning in comparative case studies (one or few treated units). With many units, averaging reduces noise}.
    \begin{enumerate}
        \item No - the series is not highly volatile
        \item Yes - but filtered or smoothed prior to applying SC
        \item Yes - fitting is performed directly on volatile data
        \item Unclear or difficult to assess based on the information provided
    \end{enumerate}

    \item \textit{Which of the following are used to construct the synthetic control and assess pre-treatment balance?}
    \begin{enumerate}
        \item Pre-treatment outcome variables (lagged outcome levels or trends)
        \item Standard demographic or economic covariates (e.g., age, gender in an individual-level study;  population, education rate, unemployment in a mezzo or macro level study)
        \item Covariates argued in the paper to be highly relevant for treatment or outcomes (e.g., think of ability or expectations proxies in an individual study; or industry composition, political alignment, fiscal variables in a macro study)
        \item Other
    \end{enumerate}

    \item \textit{How well does the synthetic control match the treated unit(s) on the outcome in the pre-treatment period? Base your answer on reported fit statistics and/or graphical comparisons of treated vs. synthetic outcomes before treatment}
    \begin{enumerate}
        \item Good fit - small and not systematic differences between treated and synthetic outcomes, visually good match and/or low RMSPE reported
        \item Poor fit - large systematic gaps between treated and synthetic outcomes before treatment
        \item Unclear or difficult to assess
        \item No quantitative or visual assessment provided
        \item Other
    \end{enumerate}

    \item \textit{Is there any donor pool trimming performed prior to the construction of the synthetic control?}
    \begin{enumerate}
        \item Yes - donor pool units were excluded due to exposure to similar policies or contemporaneous shocks
        \item Yes - donor pool units considered substantially different compared to the treated units (i.e., with the goal of avoiding interpolation bias)
        \item No - the donor pool was not trimmed or restricted
        \item Unclear or difficult to assess based on the information provided
        \item Other
    \end{enumerate}

    \item \textit{Which approaches are used to construct the synthetic control and estimate treatment effects?}
    \begin{enumerate}
        \item Classical SCM - Abadie \& Gardeazabal (2003); Abadie, Diamond \& Hainmueller (2010)
        \item Augmented or Penalized Synthetic Control - e.g., Ben-Michael, Feller \& Rothstein (2021); Abadie \& L’Hour (2021)
        \item Synthetic Difference-in-Differences (SDID)
        \item Generalized SCM (Xu, 2017)
        \item Bayesian Synthetic Control — e.g., Scott \& Varian, 2014; Hazlett \& Xu, 2018
        \item Other
    \end{enumerate}

    \item \textit{What inference methods are used to assess statistical significance and uncertainty?}
    \begin{enumerate}
        \item Permutation / Placebo tests
        \item Jackknife (e.g., leave-one-out methods)
        \item Bootstrap
        \item Conformal Inference (Chernozhukov, Wüthrich \& Zhu, 2021)
        \item Other 
    \end{enumerate}

    \item \textit{Is clustering appropriately accounted for in the inference procedure? Note: The recommended practice is to account for clustering at the level of unit assignment when the unit of analysis is at a lower level than the cluster (e.g., schools assigned but data is at student level). However, if the unit of analysis is at the level of clustering (e.g., schools assigned, and data is aggregated at school level), clustering is not necessary. Clustering is also necessary when there are multiple observations per unit used in the analysis (e.g., repeated measures) Extra: clustering is also necessary when there is a risk of interference or spillovers, but this is rarely done, so if the paper reports such, please select  ``g) Other'', ignoring all other response options}
    \begin{enumerate}
        \item Yes - clustering at level of treatment assignment
        \item Yes - clustering below treatment assignment (e.g., at class level when schools are assigned)
        \item Yes - clustering above treatment assignment (e.g., at school level, when classes are assigned)
        \item No - clustering not used and not necessary (e.g., individual-level or clustered assignment but the unit of analysis matches the treatment assignment level)
        \item No - clustering not used, but it should have been
        \item Unclear / not reported
        \item Other
    \end{enumerate}

    \item \textit{Does the study include falsification (placebo) tests? Please select the ``e) Other'' option only if it is a highly relevant falsification tests for the identification strategy}
    \begin{enumerate}
        \item Post-treatment placebo outcomes test - applying the same design / analysis replacing the outcome variable with one which should not be impacted by the treatment
        \item Placebo treatment - applying the same design / analysis replacing the treatment variable with one that should not affect the outcome. The usual approach is backdating or \enquote{in time placebo tests} - setting a placebo treatment date prior to the actual treatment date and checking if the trend diverges prior to the actual treatment
        \item Placebo population test - applying the same design / analysis in a sample where the treatment is expected to not have an effect. Consider only analysis other than the permutation tests used for inference.
        \item None of the above
        \item Other
    \end{enumerate}

    \item \textit{What type of sensitivity analyses does the study conduct? Please select the \enquote{g) Other} option only in the case of highly relevant sensitivity tests for the identification strategy}
    \begin{enumerate}
        \item Sensitivity to the violation of the identification assumptions (e.g., bounding, partial identification)
        \item Robustness to the choice of covariates (e.g., as suggested by Ferman, Pinto, and Possebom, 2020)
        \item Robustness to the choice of units in the donor pool - are results robust to excluding units from the donor pool, usually units which receive higher weights in the main analysis
        \item Robustness to alternative estimators (e.g., a bias correction approach, synthetic DID, matrix completion)
        \item Sensitivity to using alternative inference approaches (e.g., different inference method, level of clustering etc.)
        \item None of the above
        \item Other
    \end{enumerate}
\end{enumerate}

\end{document}